# Topological Data Mapping of Online Hate Speech, Misinformation, and General Mental Health: A Large Language Model Based Study

Andrew Alexander & Hongbin Wang

School of Medicine, Texas A&M University

## Abstract

The advent of social media has led to an increased concern over its potential to propagate hate speech and misinformation, which, in addition to contributing to prejudice and discrimination, has been suspected of playing a role in increasing social violence and crimes in the United States. While literature has shown the existence of an association between posting hate speech and misinformation online and certain personality traits of posters, the general relationship and relevance of online hate speech/misinformation in the context of overall psychological wellbeing of posters remain elusive. One difficulty lies in the lack of adequate data analytics tools capable of adequately analyzing the massive amount of social media posts to uncover the underlying hidden links. Recent progresses in machine learning and large language models such as ChatGPT have made such an analysis possible. In this study, we collected thousands of posts from carefully selected communities on the social media site Reddit. We then utilized OpenAI's GPT3 to derive embeddings of these posts, which are high-dimensional real-numbered vectors that presumably represent the hidden semantics of posts. We then performed various machine-learning classifications based on these embeddings in order to understand the role of hate speech/misinformation in various communities. Finally, a topological data analysis (TDA) was applied to the embeddings to obtain a visual map connecting online hate speech, misinformation, various psychiatric disorders, and general mental health.

## Introduction

The advent of social media has led to an increased concern over the potential for social media forums to contribute to hate speech, defined in the literature as "the use of offensive language, focused on a specific group of people who share a common property" (1). Online hate speech contributes to prejudice and discrimination in society when offline, and on a more severe level, social media's allowances for hate speech have led to companies such as Facebook being accused by the United Nations of contributing to genocide (2,3). Hate speech on social media has also been suspected of playing a role in social violence and crimes in the United States. A positive correlation between violent political rhetoric and mass shootings has been established (4).

Likewise, social media has proven to be a potent vector for misinformation. Misinformation, defined as information contrary to the consensus of the scientific community, has been considered a contributing factor to hate speech on social media (5, 6). Additionally, it has contributed to confusion and distrust in several areas of medicine, including vaccines, drugs, pandemics, eating disorders, and medical treatments, yielding another vector by which it may negatively impact society (6).

There have been studies investigating the potential for relationships between hate speech/misinformation and the Dark Triad personality traits, with positive correlations being established (7, 8). The Dark Triad personality traits of Narcissism, Machiavellianism, and Psychopathy have a substantial degree of overlap of characteristics with the Cluster B personality disorders defined by the *Diagnostic and Statistical Manual of Mental Disorders 5th Edition* (*DSM-5)*, suggesting the potential that these disorders could have a connection to hate speech/misinformation as well (9, 10).

The general relationship of online hate speech and misinformation with psychiatric disorders and/or the overall mental health of individuals remains elusive, however. While the Dark Triad personality traits are substantially correlated to the Cluster B personality disorders, they are not necessarily the exact same constructs (9). Therefore, they may ultimately have different relations to hate speech and misinformation. Furthermore, the Cluster B personality disorders cover only a small portion of the disorders listed in the *DSM-5*, meaning there could be additional psychiatric disorders that may have a connection with hate speech or misinformation (10). Finally, how are hate speech and misinformation related to the general mental state and psychological wellbeing of individual posters? If it were possible to identify hidden psychiatric disorders or traits of them with misinformation or hate speech in the overall context of mental health, the gained knowledge could be used to implement more general and effective counter strategies against hate speech or misinformation.

Recent advances in artificial intelligence (AI) and machine learning (ML) have provided new tools to tackle these problems (11, 12, 13). In particular, large language models (LLMs), as exemplified by GPT3 and ChatGPT (14,15,16) have been demonstrated to be especially powerful in understanding natural languages and performing various natural language processing tasks. One of key features of LLMs is to learn, via massively pre-training a transformer neural network (17,18) with a context-dependent representation, called an embedding, of any text input. In the case of GPT3, such an embedding is a 1536-dimensional real-valued vector. These embeddings approximate a computer-friendly representation of the semantics of the text input, and therefore can be used for downstream tasks such as text classification, summarization and question-answering. For example, in so-called zero-shot classification (19), a LLM can be used to classify any input by simply calculating and comparing the vector distance of the input and a pre-specified class label.

In this work, we intend to utilize LLMs to study the relevance of online hate speech and misinformation in the context of the general mental health of posters. In particular, we collect thousands of online posts from a carefully selected group of communities in Reddit. We then derive the embeddings of these posts using GPT3. We explore the correlation of hate and misinformation with specific psychiatric disorders using zero-shot classification. Finally, to gain a deeper insight on hate speech and misinformation in the general context of mental health, we carry out a topological data analysis (TDA) on these embeddings to obtain an overarching topological mapping of the dataset (20, 21, 22). The map provides a snapshot of the shape of the mental health space and relationships between psychiatric disorders, hate speech, and misinformation within that space.

## Methodology

*Social Media Selection*

We opted to utilize the social media site Reddit as our source to draw embeddings from. Reddit has had a substantial number of communities associated with misinformation and hate speech in its past and present, making it an excellent repository of this data for our purposes (23). Additionally, Reddit is split into "subreddits", which are effectively communities dedicated to a specific topic (24). This is beneficial to our study, as it allows for the existence of communities dedicated to specific psychiatric disorders, which would generally self-select for members with the corresponding disorder. Furthermore, Reddit has the advantage of anonymity; Reddit usernames are not tied to an individual's personal name, and identifying information is only divulged on Reddit by a user if they wish to divulge it (24). This is of substantial value when collecting data on psychiatric disorders, where stigmatization could prevent individuals from open discussion if anonymity was unavailable (25).

*Community Selection*

A total of 54 communities were selected that could bear relevance to four different classes: Hate Speech, Misinformation, Psychiatric Disorders (which had subclasses for each psychiatric disorder represented by the communities we obtained embeddings from), and Control (Table 1). Hate speech and misinformation communities on Reddit were selected based on either their designation as such in prior literature or records of ban or quarantine by Reddit administration for reasons pertaining to hate, violence, or misinformation (26-42). If a community could be considered as spreading both hate speech and misinformation, we categorized it by personal assessment as to which was the more dominant theme of the community. Communities identified as representing a psychiatric disorder were primarily based on their own declaration as such and personal assessment of posts to ascertain that this was the genuine goal of the community. Communities were considered to be control communities if they did not meet the criteria for any of the other classes.

*Embedding Types: Individual User Post Embeddings and Distilled Embeddings*

We distinguished two types of embeddings: "Individual User Post" (IUP) embeddings, created based on each individual post, and "distilled" embeddings, created based on an aggregate of posts from multiple users within a single community. The reason to derive distilled embeddings is to obtain an averaged representation of a community that dilutes the influence of an individual's unique speech patterns. The unique speech patterns of the individual users, being unique, would fail to stand out after aggregation, whereas the characteristic speech patterns for a community would become more influential due to their presence in a large number of users. In this sense, we are attempting to create embeddings that are more representative of psychiatric disorder and hate speech/misinformation communities than IUP embeddings could provide.

**Table 1**. Reddit Communities Used for Embedding Creation

| Subreddit | Number of Distilled Embeddings | IUP Embeddings | Classification | Justification for Classification |
|---|---|---|---|---|
| r/ADHD | 31 | Yes | ADHD | Self-Declared Psychiatric Disorder |
| r/adhdwomen | 28 | No | ADHD | Self-Declared Psychiatric Disorder |
| r/depression | 29 | Yes | Depression | Self-Declared Psychiatric Disorder |
| r/depressed | 30 | No | Depression | Self-Declared Psychiatric Disorder |
| r/depressionregimen | 28 | No | Depression | Self-Declared Psychiatric Disorder |
| r/bpd | 33 | Yes | Borderline Personality Disorder (BPD) | Self-Declared Psychiatric Disorder |
| r/BorderlinePDisorder | 25 | No | Borderline Personality Disorder (BPD) | Self-Declared Psychiatric Disorder |
| r/AnorexiaNervosa | 20 | No | Eating Disorders | Self-Declared Psychiatric Disorder |
| r/BingeEatingDisorder | 23 | No | Eating Disorders | Self-Declared Psychiatric Disorder |
| r/bulimia | 19 | Yes | Eating Disorders | Self-Declared Psychiatric Disorder |
| r/narcissism | 29 | Yes | Narcissistic Personality Disorder (NPD) | Self-Declared Psychiatric Disorder |
| r/NPD | 25 | No | Narcissistic Personality Disorder (NPD) | Self-Declared Psychiatric Disorder |
| r/aspd | 16 | Yes | Antisocial Personality Disorder (ASPD) | Self-Declared Psychiatric Disorder |
| r/alcoholism | 26 | Yes | Substance Use Disorder | Self-Declared Psychiatric Disorder |
| r/addiction | 28 | No | Substance Use Disorder | Self-Declared Psychiatric Disorder |
| r/alcoholicsanonymous | 23 | No | Substance Use Disorder | Self-Declared Psychiatric Disorder |
| r/cripplingalcoholism | 34 | No | Substance Use Disorder | Self-Declared Psychiatric Disorder |
| r/bipolar2 | 20 | No | Bipolar Disorder | Self-Declared Psychiatric Disorder |
| r/BipolarReddit | 24 | No | Bipolar Disorder | Self-Declared Psychiatric Disorder |
| r/bipolar | 21 | Yes | Bipolar Disorder | Self-Declared Psychiatric Disorder |
| r/autism | 21 | Yes | Autism | Self-Declared Psychiatric Disorder |
| r/aspergers | 28 | No | Autism | Self-Declared Psychiatric Disorder |
| r/Anxiety | 24 | Yes | Anxiety | Self-Declared Psychiatric Disorder |
| r/Agoraphobia | 27 | No | Anxiety | Self-Declared Psychiatric Disorder |
| r/Anxietyhelp | 18 | No | Anxiety | Self-Declared Psychiatric Disorder |
| r/OCD | 26 | Yes | Obsessive Compulsive Disorder (OCD) | Self-Declared Psychiatric Disorder |
| r/ptsd | 33 | Yes | Post-Traumatic Stress Disorder (PTSD) | Self-Declared Psychiatric Disorder |
| r/CPTSD | 38 | Yes | Complex Post-Traumatic Stress Disorder (CPTSD) | Self-Declared Psychiatric Disorder |
| r/Suicidal_Thoughts | 21 | No | Suicidality | Self-Declared Psychiatric Disorder |
| r/SuicideWatch | 26 | Yes | Suicidality | Self-Declared Psychiatric Disorder |
| r/schizoaffective | 20 | Yes | Schizophrenia | Self-Declared Psychiatric Disorder |
| r/schizophrenia | 17 | Yes | Schizophrenia | Self-Declared Psychiatric Disorder |
| r/Schizotypal | 27 | Yes | Schizotypal Personality Disorder | Self-Declared Psychiatric Disorder |
| r/Schizoid | 28 | Yes | Schizoid Personality Disorder | Self-Declared Psychiatric Disorder |
| r/NoNewNormal | 10 | No | Misinformation | Misinformation community dedicated to COVID-19 conspiracies; banned (26) |
| r/ivermectin | 10 | No | Misinformation | Misinformation community dedicated to the use of Ivermectin to treat COVID-19; quarantined (27) |
| r/vaccinelonghaulers | 21 | No | Misinformation | Misinformation community dedicated to misinformation about vaccine side effects; quarantined (28) |
| r/conspiracy | 9 | No | Misinformation | Largest conspiracy community on Reddit (29) |
| r/greatawakening | 15 | No | Misinformation | QAnon conspiracy community; banned (30) |
| r/MGTOW | 11 | No | Hate Speech | Banned for hate speech (31,32) |
| r/Incels | 10 | No | Hate Speech | Banned for promoting violence against women (33,34) |
| r/TruFemcels | 23 | No | Hate Speech | Banned for hate speech (35) |
| r/Gender_Critical | 10 | No | Hate Speech | Banned for hate speech (36) |
| r/KotakuInAction | 11 | No | Hate Speech | Several publications regarding racism and sexism (37, 38) |
| r/MensRights | 21 | No | Hate Speech | Several citations regarding misogyny (39,40) |
| r/TheRedPill | 124 | No | Hate Speech | Several citations regarding misogyny (39,40) |
| r/CringeAnarchy | 4 | No | Hate Speech | Banned for violent content (41) |
| r/Chodi | 4 | No | Hate Speech | Banned for hate speech (42) |
| r/Teenagers | 6 | Yes | Control | No apparent ties to hate speech, misinformation, or mental health |
| r/ShowerThoughts | 4 | Yes | Control | No apparent ties to hate speech, misinformation, or mental health |
| r/apple | 11 | Yes | Control | No apparent ties to hate speech, misinformation, or mental health |
| r/ApplyingToCollege | 16 | Yes | Control | No apparent ties to hate speech, misinformation, or mental health |
| r/Agriculture | 6 | Yes | Control | No apparent ties to hate speech, misinformation, or mental health |
| r/askscience | 12 | Yes | Control | No apparent ties to hate speech, misinformation, or mental health |

*Embedding Generation*

A total of 1000 posts were pulled for each of the 54 communities selected from the "*Subreddit comments/submissions 2005-06 to 2022-12*" academic torrent (43). Specifically, we pulled posts starting from September 15, 2022 and working backwards until 1000 posts had been collected. The text data used for each post was comprised of the text used for the title, and the text used inside the post itself from the author.

Individual posts from a community were combined to derive distilled embeddings. The hate speech/misinformation communities received distilled embeddings, but not IUP embeddings, as these two embedding classes were only used later in the zero-shot classification analyses, where they served as testing data comprised entirely of distilled embeddings. For psychiatric disorder communities, we opted to obtain distilled embeddings of multiple communities as opposed to one community per disorder whenever possible. The rationale behind this was threefold: First, there is always the chance that a community claiming to be representative of a disorder is not actually representative of it. Second, a community could develop its own jargon and speech patterns that would be independent of the psychiatric disorder itself. Bringing in additional communities that may have differing community-specific characteristics could potentially mitigate this effect. While we created IUP embeddings of psychiatric disorder communities, we did not apply this multiple-community treatment to IUP embeddings of psychiatric disorders, as we considered them to already be inherently assigning extra weight to speech patterns that were unique to individuals (and therefore rendering other attempts to erase the influence of unique speech patterns to be rather ineffective).

OpenAI's GPT3 embedding model (text-embedding-ada-002) was used for the generation of embeddings (44). All embeddings were generated as a vector space with 1536 dimensions. Given the current limitation of GPT3 on maximum allowed token length (<8192), multiple distilled embeddings were possible for a community. Efforts were taken to make sure an individual post was not split across different community embeddings. As different communities and posts have differing complexities in their text, and therefore different token counts, this led to communities having different numbers of distilled embeddings (Table 1). For IUP embeddings, a set of 50 posts each were collected from the selected communities, working backwards from September 15, 2022 like before.

*Zero-shot Classification*

Four zero-shot classification tasks were performed to explore the interplay of different embedding types and hate speech/misinformation with disorders (Table 2). They are called zero-shot classification tasks due to the fact that the model is trained based on one dataset and then tested directly on a separate dataset for different labelling.

Classification tasks 1 and 2 were performed with the intent of assessing the best combination of IUP and distilled embeddings to use when performing zero-shot classification of hate speech/misinformation. Specifically, classification task 1 was a classification of psychiatric disorder embeddings using IUP embeddings as training data and distilled embeddings as testing data, and vice versa for classification task 2.

Based on the results of classification tasks 1 & 2, classification tasks 3 & 4 were set as zero-shot classifications of the distilled embeddings of a total of 14 hate speech and misinformation communities. With zero-shot classification, the model is not given any embeddings from hate speech or misinformation communities as training data, and these embeddings comprise the entirety of the model's testing data; as such, it is forced to classify hate speech and misinformation embeddings as being closer to either a psychiatric disorder distilled embedding or a control distilled embedding. Communities that appeared to be solely dedicated to misinformation instead of a combination of misinformation and hate speech were classified separately from hate speech communities, due to concerns that classifications for psychiatric disorders could differ between the two.

**Table 2**. Zero-shot classification tasks

| Classification Task | Training Data | Testing Data |
|---|---|---|
| 1 | All psychiatric disorder and control IUP embeddings | All psychiatric disorder and control distilled embeddings |
| 2 | All psychiatric disorder and control distilled embeddings | All psychiatric disorder and control IUP embeddings |
| 3 | All psychiatric disorder and control IUP embeddings | All hate and control distilled embeddings |
| 4 | All psychiatric disorder and control IUP embeddings | All misinformation and control distilled embeddings |

*Topological Data Analysis*

Text embeddings derived from LLMs are usually very high-dimensional (1536 dimensions in our case). While such high dimensionality is useful to capture complex hidden semantics of any given text input, it also renders traditional statistical methods (e.g., multivariant linear regression) and data visualization less effective. To further pinpoint the relation between hate speech/misinformation and psychiatric disorders in the general context of mental health, we performed a TDA to explore the global shape of derived embeddings. In general, TDA is an unsupervised machine learning method that excels in handling high-dimensional data (20,21,22). Unlike linear models such as principal component analysis, which have difficulty preserving the shape of high-dimensional datasets, TDA can capture nonlinearity by projecting a high-dimensional dataset onto a lower dimensional space in a systematical and sensible way. For example, in Mapper (45), a high-dimensional dataset is first projected into a lower dimension space by a filter function. In the lower dimension, the data is then segmented into several overlapping sections, called covers. Clustering is then performed in each cover, and the resultant clusters (referred to as nodes) that share data points are then connected by edges. By doing so, Mapper is capable of producing a visual representation of the global shape of the data without sacrificing non-linearity.

We used Mapper Interactive to generate a TDA mapping of the distilled embeddings (46). First, a principal component analysis was carried out and the first two principal components were extracted to serve as a 2-dimensional filter function for Mapper. Both dimensions used 10 covers, each with 50% overlap. For our clustering algorithm, we used DBSCAN at 0.5 eps and 2 minimum samples.

## Results

*Classifications 1 and 2: Cross-Comparison of Individual User Post and Distilled Embeddings as Training and Testing Data*

Here we sought to evaluate how a model trained on IUP embeddings would classify distilled embeddings, and vice versa. Evaluation of the psychiatric disorder embeddings showed excellent accuracy when IUP embeddings were used for training (86% accuracy), whereas distilled embeddings showed difficulty in adequately classifying individual posts, with an accuracy of 45% (Tables 3-4, Figures 1-2). Inspection showed that many of the inaccuracies seen in classification of IUP embeddings by a model trained on distilled embeddings were attributable to a frequent misclassification of embeddings as substance use disorders.

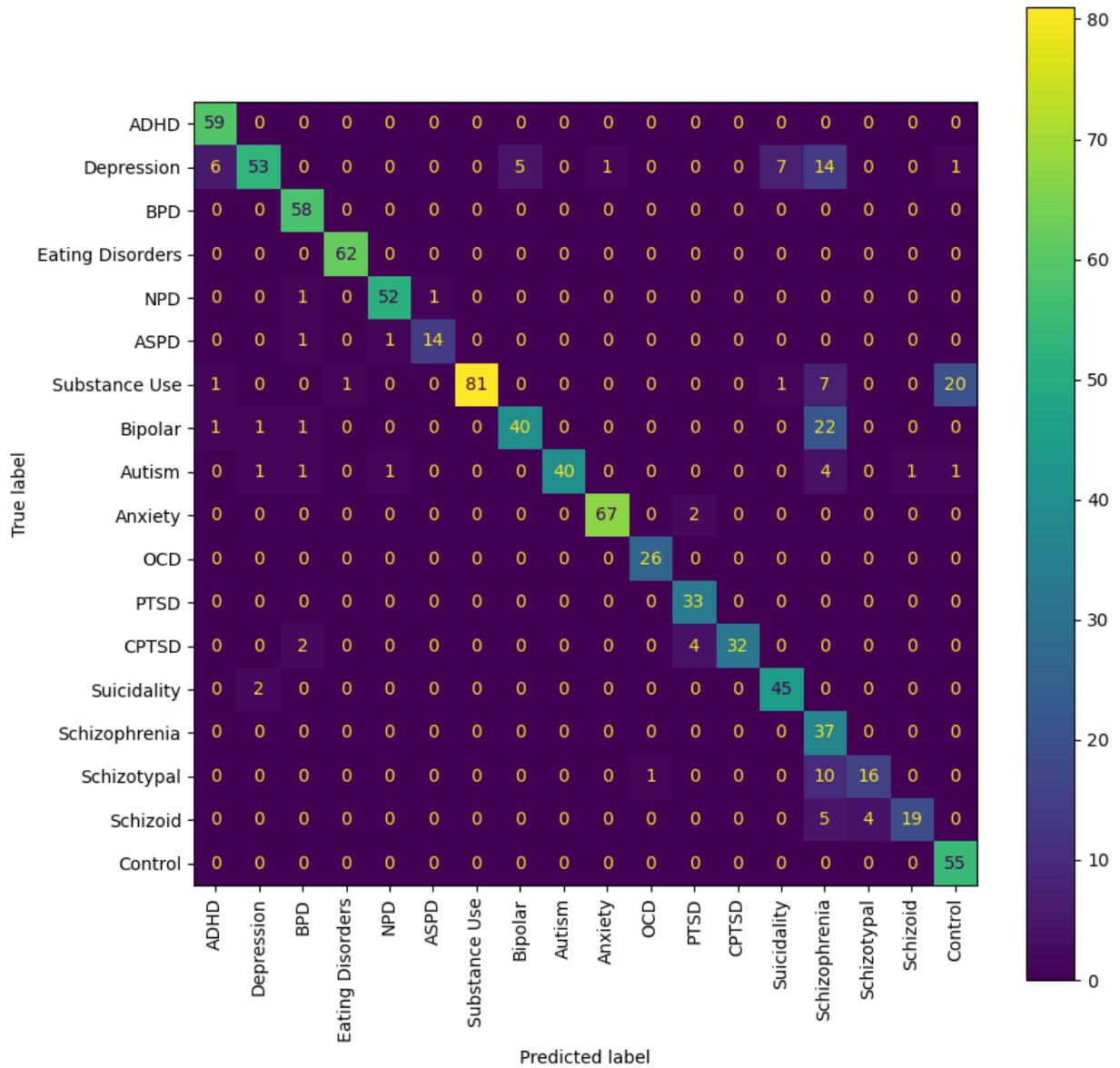

**Figure 1**. Confusion Matrix of Distilled Embeddings Classified Using Model Trained with Individual User Post Embeddings.

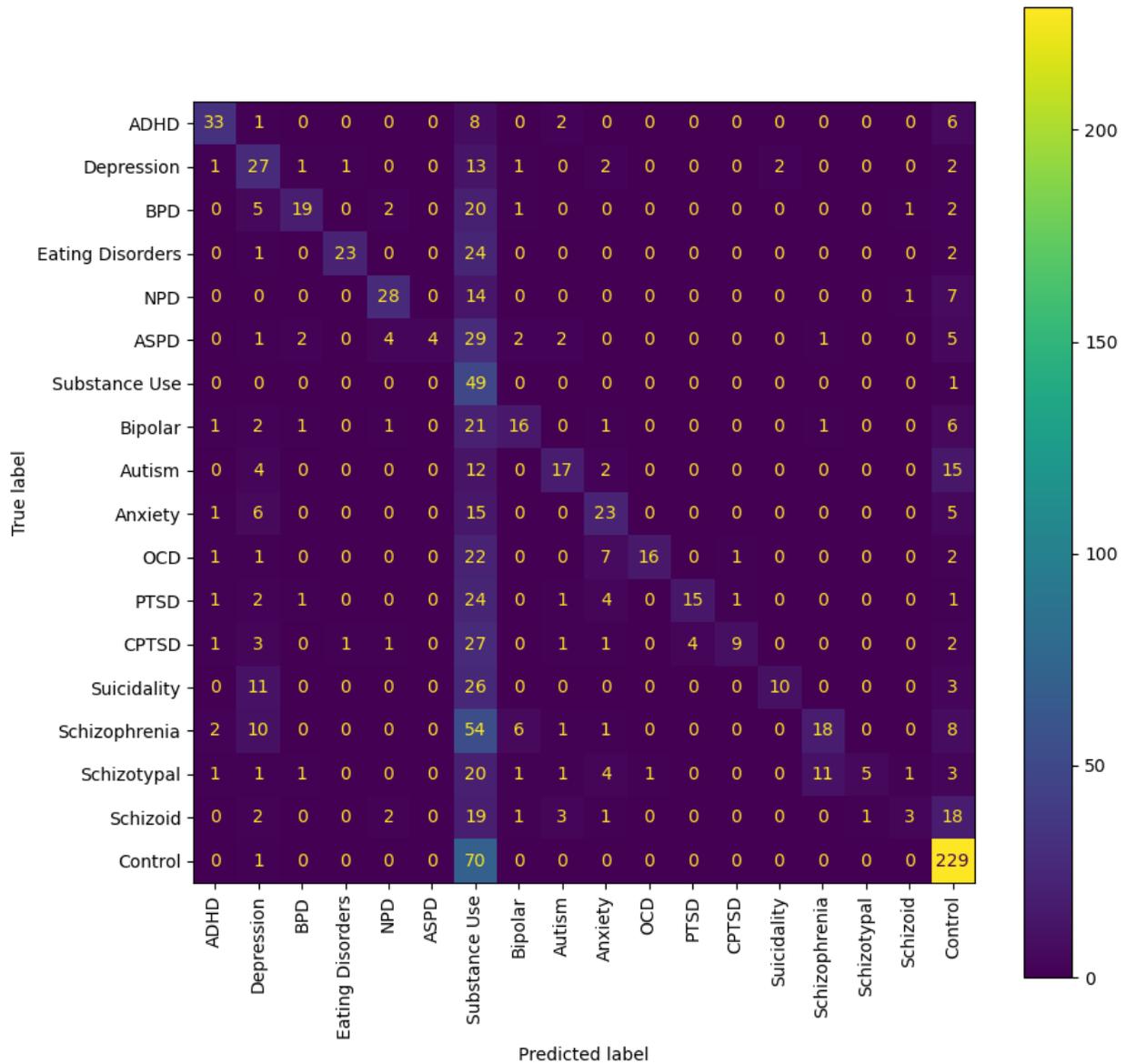

**Figure 2**. Confusion Matrix of Individual User Post Embeddings Classified Using Model Trained with Distilled Post Embeddings.

**Table 3**. Classification Report of Distilled Embeddings Classified Using Model Trained with Individual User Post Embeddings.

| Category | Precision | Recall | f1-Score | Support |
|---|---|---|---|---|
| ADHD | 0.88 | 1.00 | 0.94 | 59 |
| Depression | 0.93 | 0.61 | 0.74 | 87 |
| Borderline Personality Disorder | 0.91 | 1.00 | 0.95 | 58 |

| Category | Precision | Recall | f1-Score | Support |
| --- | --- | --- | --- | --- |
| Eating Disorders | 0.98 | 1.00 | 0.99 | 62 |
| Narcissistic Personality Disorder | 0.96 | 0.96 | 0.96 | 54 |
| Antisocial Personality Disorder | 0.93 | 0.88 | 0.90 | 16 |
| Substance Use Disorders | 1.00 | 0.73 | 0.84 | 111 |
| Bipolar Disorder | 0.89 | 0.62 | 0.73 | 65 |
| Autism Spectrum Disorder | 1.00 | 0.82 | 0.90 | 49 |
| Anxiety Disorders | 0.99 | 0.97 | 0.98 | 69 |
| Obsessive-Compulsive Disorder | 0.96 | 1.00 | 0.98 | 26 |
| Post-Traumatic Stress Disorder | 0.85 | 1.00 | 0.92 | 33 |
| Complex Post-Traumatic Stress Disorder | 1.00 | 0.84 | 0.91 | 38 |
| Suicidality | 0.85 | 0.96 | 0.90 | 47 |
| Schizophrenia/Schizoaffective | 0.37 | 1.00 | 0.54 | 37 |
| Schizotypal | 0.80 | 0.59 | 0.68 | 27 |
| Schizoid | 0.95 | 0.68 | 0.79 | 28 |
| Control | 0.71 | 1.00 | 0.83 | 55 |
| | | | | |
| Accuracy | | | 0.86 | 921 |
| Macro Average | 0.89 | 0.87 | 0.86 | 921 |
| Weighted Average | 0.90 | 0.86 | 0.86 | 921 |

**Table 4**. Classification Report of Individual User Post Embeddings Classified Using Model Trained with Distilled Post Embeddings.

| Category | Precision | Recall | f1-Score | Support |
| --- | --- | --- | --- | --- |
| ADHD | 0.79 | 0.66 | 0.72 | 50 |
| Depression | 0.35 | 0.54 | 0.42 | 50 |
| Borderline Personality Disorder | 0.76 | 0.38 | 0.51 | 50 |
| Eating Disorders | 0.92 | 0.46 | 0.61 | 50 |
| Narcissistic Personality Disorder | 0.74 | 0.56 | 0.64 | 50 |
| Antisocial Personality Disorder | 1.00 | 0.08 | 0.15 | 50 |
| Substance Use Disorders | 0.10 | 0.98 | 0.19 | 50 |
| Bipolar Disorder | 0.57 | 0.32 | 0.41 | 50 |
| Autism Spectrum Disorder | 0.61 | 0.34 | 0.44 | 50 |
| Anxiety Disorders | 0.50 | 0.46 | 0.48 | 50 |
| Obsessive-Compulsive Disorder | 0.94 | 0.32 | 0.48 | 50 |
| Post-Traumatic Stress Disorder | 0.79 | 0.30 | 0.43 | 50 |
| Complex Post-Traumatic Stress Disorder | 0.82 | 0.18 | 0.30 | 50 |
| Suicidality | 0.83 | 0.20 | 0.32 | 50 |
| Schizophrenia/Schizoaffective | 0.58 | 0.18 | 0.27 | 100 |
| Schizotypal | 0.83 | 0.10 | 0.18 | 50 |
| Schizoid | 0.50 | 0.06 | 0.11 | 50 |
| Control | 0.72 | 0.76 | 0.74 | 300 |
| | | | | |
| Accuracy | | | 0.45 | 1200 |
| Macro Average | 0.69 | 0.38 | 0.41 | 1200 |
| Weighted Average | 0.69 | 0.45 | 0.47 | 1200 |

Several inaccuracies appear to be present in the results obtained from classifying distilled embeddings with a model trained by IUP embeddings. The three disorder community types with the greatest degree of misclassification were depression, CPTSD, and Schizoaffective Disorder/Schizophrenia. However, the existence of comorbidities suggests that several of these inaccuracies may very well not be inaccuracies at all, or are at the least very close to a correct classification. The depression distilled embedding misclassifications were in four classes: ADHD, Bipolar Disorder, Anxiety, and Suicidality, all of which are comorbid with depression, or in the case of Bipolar Disorder, contain depressive episodes within their diagnostic criteria (10, 47, 48).

CPTSD's misclassifications were in PTSD and Borderline Personality Disorder. There is a high degree of feature overlap between BPD and CPTSD, and the debate over whether CPTSD is distinct enough from PTSD to warrant its own diagnosis in the DSM-5 is indicative of substantial similarities between these two disorders as well (49, 50).

Finally, Schizophrenia and Schizoaffective Disorder were misclassified as two other disorders: Bipolar Disorder and Schizotypal Personality Disorder. Both of these disorders are capable of the psychotic features present in Schizophrenia and Schizoaffective Disorder, offering a reasonable explanation for the misclassification (10, 51). Overall, the correct classifications and the misclassifications being centered around disorders that are comorbid with the target disorder or downright include it suggest that a very small dataset of IUP embeddings was sufficient to train a model that could accurately classify a distilled embedding by the disorder they represented.

*Classifications 3 and 4: Classification of Misinformation/Hate Communities by Disorder*

Based on the results of the previous analyses, we concluded that the best design for our classification of hate speech and misinformation would be to utilize IUP embeddings as our training data, and distilled embeddings as our testing data. Zero-shot classification of the distilled embeddings for hate speech communities had a split between control classifications (21.36%) and psychiatric disorder classifications (78.64%) (Figures 3-4). The four disorders that were most often classed for misinformation/hate communities were Schizoid Personality Disorder and the three Cluster B Personality Disorders included in the study: Narcissistic Personality Disorder, Borderline Personality Disorder, and Antisocial Personality Disorder. Of these disorders, Schizoid Personality Disorder accounted for 25.73% of all embeddings and 32.72% of embeddings classified as belonging to a psychiatric disorder community, while the Cluster B Personality Disorders accounted for 47.09% of all embeddings and 59.88% of embeddings classified as pertaining to a psychiatric disorder community.

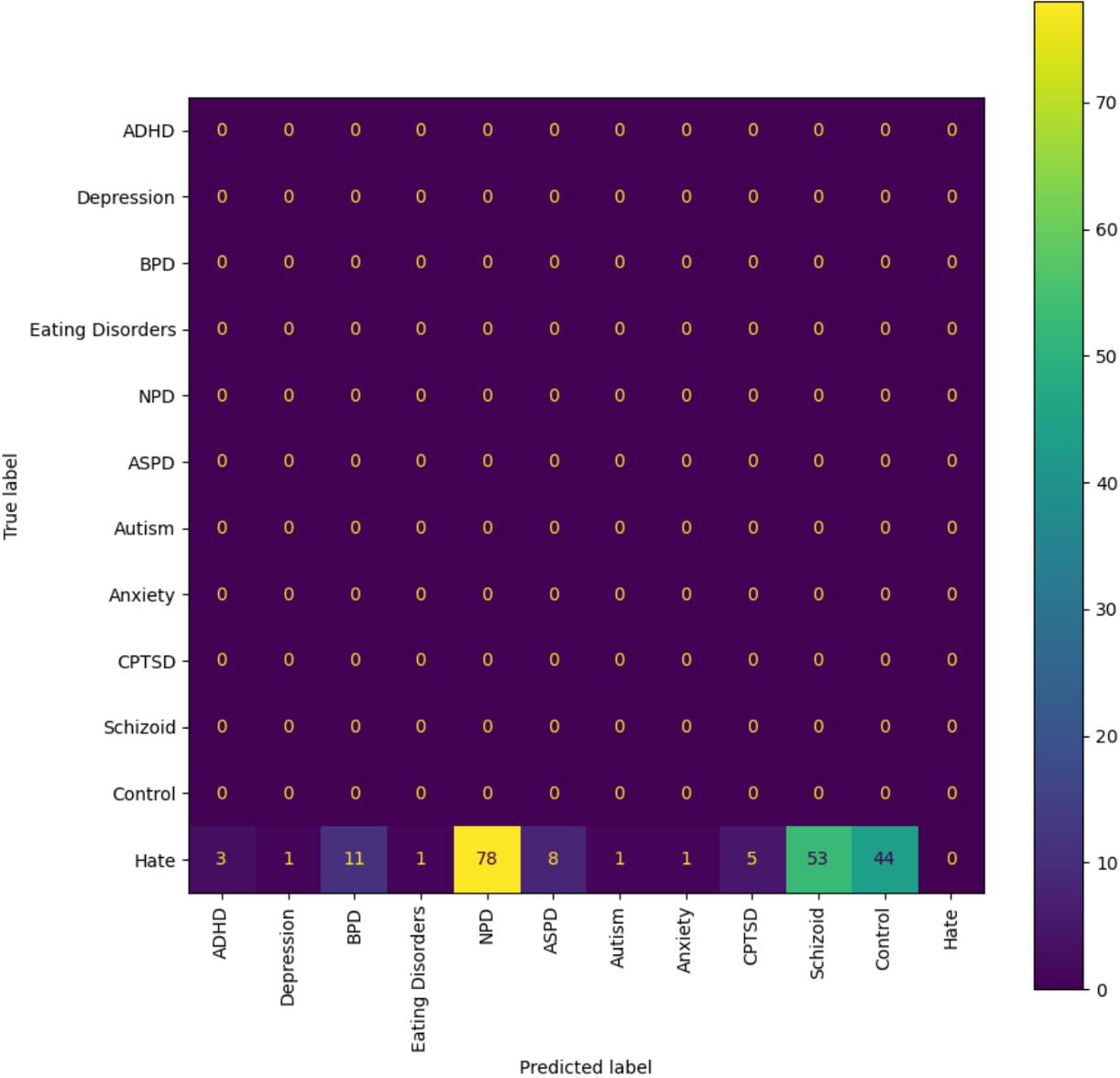

**Figure 3**. Confusion Matrix of Zero-Shot Classification of Hate Speech Distilled Embeddings Using Individual User Post Embeddings as Training Data

In communities that are more purely misinformation rather than a combination of misinformation and hate speech, the dominant psychiatric disorder classification was anxiety disorders (13.85%), though the vast majority of classifications tended to lean towards control

(73.85%), with only 26.15% of embeddings being classified as pertaining to psychiatric disorder communities.

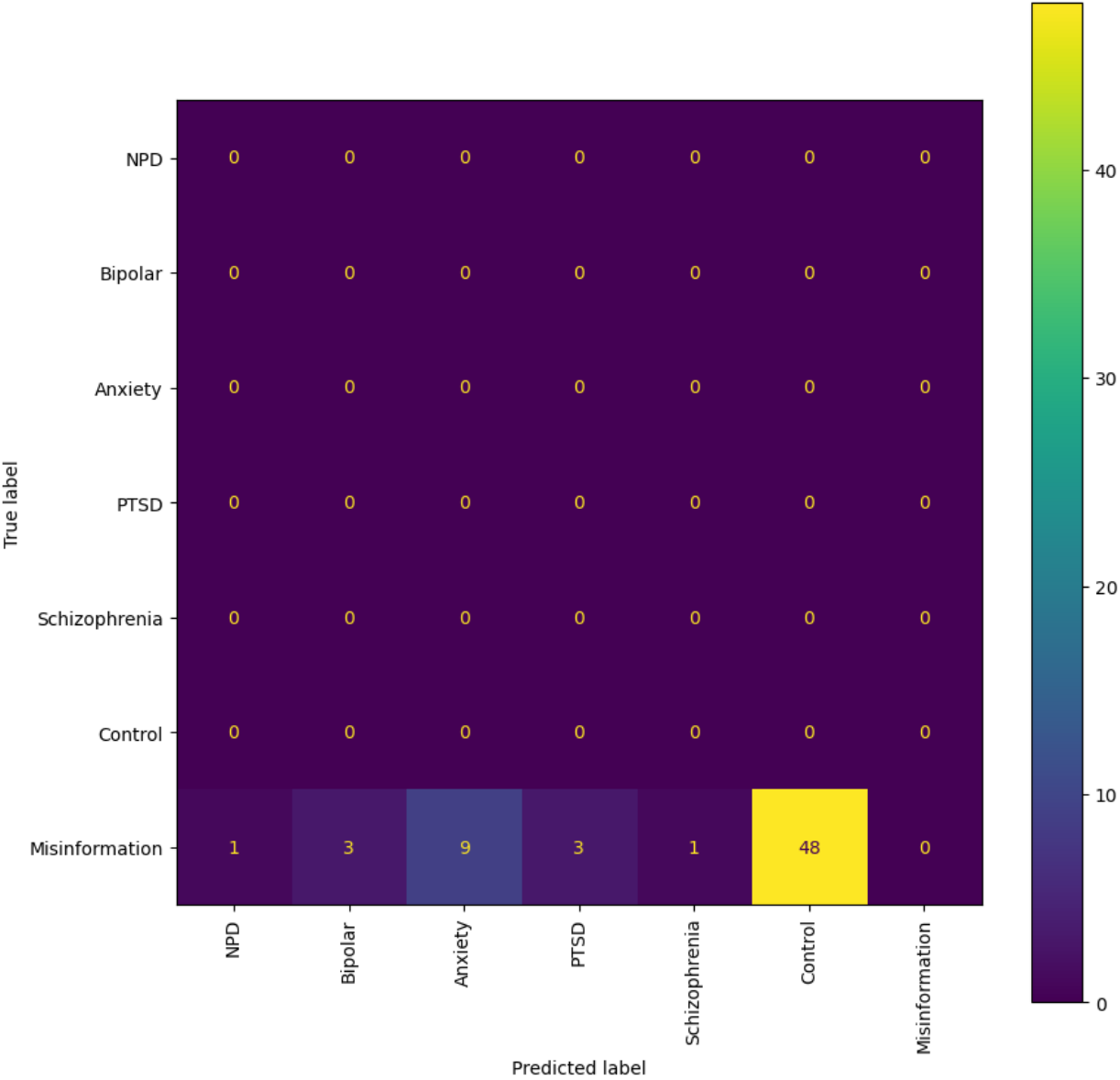

**Figure 4**. Confusion Matrix of Zero-Shot Classification of Misinformation Distilled Embeddings Using Individual User Post Embeddings as Training Data.

*Topological Data Analysis*

The topological mapping of our collection of embeddings yielded several interesting observations. Firstly, all of the psychiatric disorder embeddings had either direct or indirect edges with each other, all occupying their own portion of space within the mapping (Figure 5a). Secondly, this space comprising the psychiatric disorders was connected directly to a space comprising all of the hate speech embeddings (Figure 5b) and almost all of the misinformation embeddings (Figure 5c). What makes this notable is that almost all of the control embeddings had no connection to this portion of the space; this lack of connection indicates that this hate speech/misinformation/psychiatric disorder space shares certain characteristics within their speech patterns that are not generally seen within the control embeddings (Figure 5d).

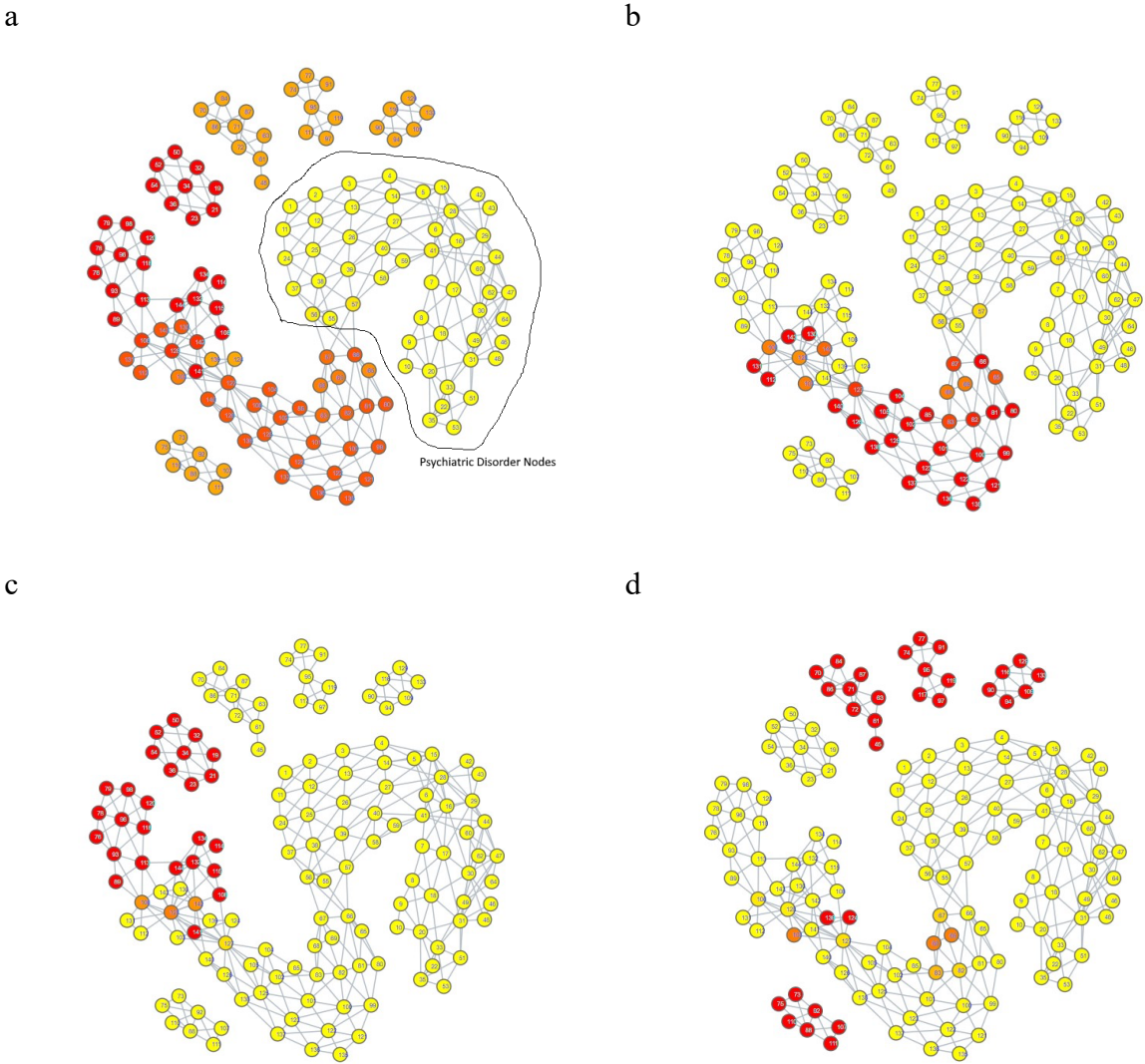

**Figure 5**. TDA Mapping of All Distilled Embeddings. (a) Colored by Category. Yellow nodes signify a node comprised of psychiatric disorder embeddings. (b) Colored by Percent Hate Speech Embeddings. Yellow = No Hate Speech Embeddings, Red = Entirely Comprised of Hate Speech Embeddings. (c) Colored by Percent Misinformation Embeddings. Yellow = No Misinformation Embeddings, Red = Entirely Comprised of Misinformation Embeddings. (d) Colored by Percent Control Embeddings. Yellow = No Control Embeddings, Red = Entirely Comprised of Control Embeddings.

While these observations are notable enough on their own, further inspection in conjunction with the results of our zero-shot classification analyses reveals that the patterns seen in the TDA mapping could be seen as predictive of the zero-shot classification results. The portion of space taken up by hate speech is closer and more immediately connected to the psychiatric disorders than the space occupied by misinformation. This corresponds to how there were far more hate speech embeddings classified as psychiatric disorders compared to misinformation embeddings; hate speech embeddings shared more similarities to psychiatric disorder embeddings. The hate speech/misinformation embeddings classified as control can be explained by the nodes in the hate speech and misinformation portions of space that contain a few control embeddings. Additionally, the portion of misinformation embeddings that occupied their own space lacked any connection with psychiatric disorders; it stands to reason that those embeddings would become classified as control.

Looking at the psychiatric disorder portion of the space in more granular detail illustrates further consistency with our zero-shot classification results. Each type of psychiatric disorder community occupied its own subsection of the space. If a psychiatric disorder community occupied a space closer to the connection point between the psychiatric disorder space and the hate speech space, then its speech patterns were more similar to those seen in hate speech. Inspection indicated that the psychiatric disorder communities occupying the connection point were the Cluster B personality disorder communities, and Schizoid Personality Disorder, with the communities for Narcissistic Personality Disorder and Schizoid Personality Disorder being most directly connected (Figure 6).

These two personality disorder communities were also the ones with the most hate speech embeddings attributed to them, lending credence to the idea that this TDA mapping is illustrating a trend where increased proximity of a psychiatric disorder community's distilled embeddings to the hate speech portion of the space suggests it has an increased likelihood of having hate speech embeddings classified to it.

Additionally, the community embeddings for r/CPTSD were prominent in this region as well, corresponding with its mild amount of hate speech embeddings classified towards it (Figure 7).

Communities for psychiatric disorders that received little to no hate speech classifications were further away from the connection point; figures for these disorders can be seen in the Supplemental.

Based on the TDA mapping, we decided to examine what the zero-shot classification results would look like if the Schizoid Personality Disorder and Narcissistic Personality Disorder communities were excluded from the training data (Supplemental Figures N-P). Doing so led to a redistribution of their assigned hate speech embeddings to CPTSD, BPD, and ASPD. An additional zero-shot classification excluding the aforementioned disorders in addition to CPTSD indicated that the embeddings assigned to CPTSD were transferred to BPD and ASPD (Supplemental Figure Q).

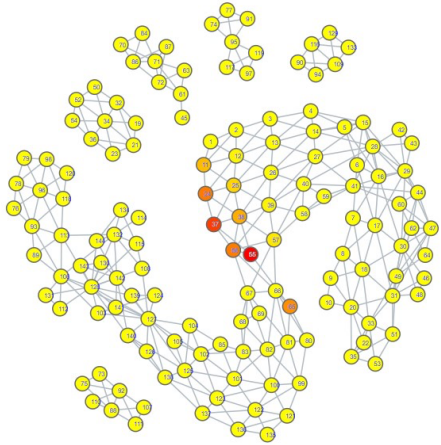 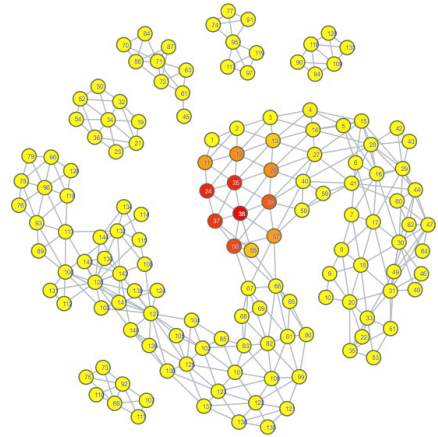

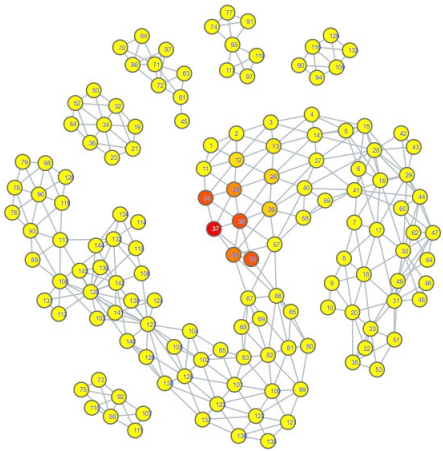 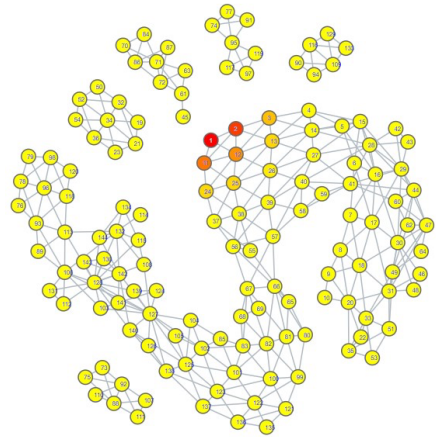

**Figure 6**. TDA Mapping of All Distilled Embeddings, Colored by Personality Disorder Embeddings Closest to Hate Speech. (a) Colored by Percent Narcissistic Personality Disorder Embeddings. Yellow = No Narcissistic Personality Disorder Embeddings, Red = Entirely Comprised of Narcissistic Personality Disorder Embeddings. (b) Colored by Percent Schizoid Personality Disorder Embeddings. Yellow = No Schizoid Personality Disorder Embeddings, Red = Entirely Comprised of Schizoid Personality Disorder Embeddings. (c) Colored by Percent Antisocial Personality Disorder Embeddings. Yellow = No Antisocial Personality Disorder Embeddings, Red = Entirely Comprised of Antisocial Personality Disorder Embeddings. (d) Colored by Percent Borderline Personality Disorder Embeddings. Yellow = No Borderline Personality Disorder Embeddings, Red = Entirely Comprised of Borderline Personality Disorder Embeddings.

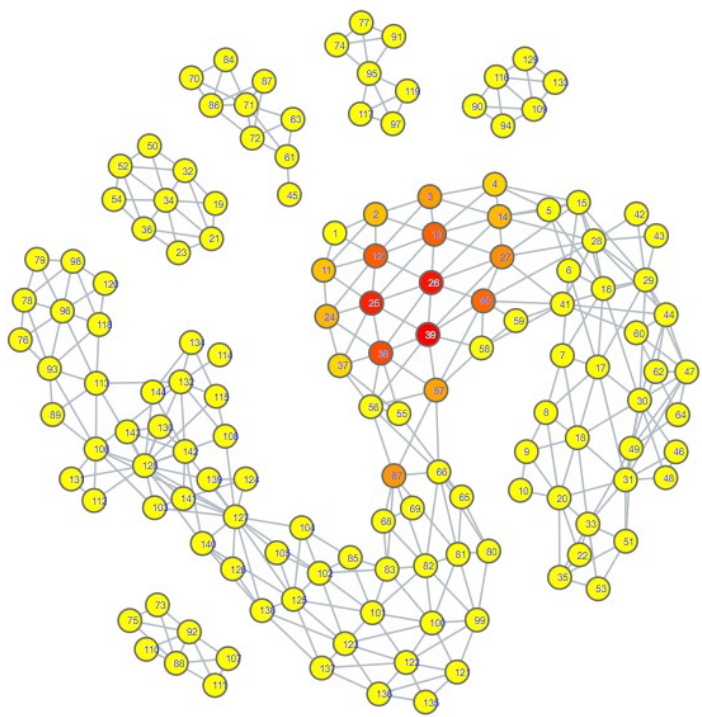

**Figure 7**. TDA Mapping of All Distilled Embeddings, Colored by Percent Complex Post-Traumatic Stress Disorder Embeddings. Yellow = No Complex Post-Traumatic Stress Disorder Embeddings, Red = Entirely Comprised of Complex Post-Traumatic Stress Disorder Embeddings.

## Discussion

Our classification model yielded an interesting classification pattern with regards to hate speech communities on Reddit: Their speech patterns are close enough to those seen in Reddit communities for Cluster B Personality Disorders that they can be classed as such under zero-shot classification. In a certain sense, this result is not surprising, as it is consistent with studies relating the Dark Triad to hate speech, and these disorders are strongly linked with a lack of empathy for others and violence against others, traits that can be seen as a product of hate speech (52-56). Although our zero-shot classification results initially appeared to indicate that the only Cluster B personality disorder of import was Narcissistic Personality Disorder, our TDA mapping indicated that CPTSD and the other Cluster B personality disorders could be considered as close to the connection point with the hate speech region as well. This suggested that they could receive the hate speech classifications as well, and that Narcissistic Personality Disorder and Schizoid Personality Disorder were merely the disorders with speech patterns most similar to hate speech, not the **only** disorders with similarity. When we excluded Narcissistic Personality Disorder and Schizoid Personality Disorder from zero-shot classification (Supplemental Figures N-P), this suspicion proved true; while some hate speech embeddings were reclassified as control, many went to the other Cluster B disorders and CPTSD. By making use of TDA, we were thus able to enhance the results of our zero-shot classification from depicting only two disorders as being substantially tied to hate speech to five disorders, those of Antisocial Personality Disorder, Narcissistic

Personality Disorder, Borderline Personality Disorder, Complex Post-Traumatic Stress Disorder, and Schizoid Personality Disorder. While this shows that TDA is able to enhance zero-shot classification, the means by which we were able to use it to do so also indicates that the mapping of the psychiatric disorders could be viewed as a spectrum of similarity to hate speech; the closer a psychiatric disorder is to the hate speech region of space, the more similar it is, and vice versa.

The presence of Schizoid Personality Disorder as the second-largest classification for hate speech is surprising. Schizoid Personality Disorder's restricted expression of emotions and detachment from social relationships does not appear conducive to hate speech, which would likely be generated from emotionality and from considering oneself as part of an in-group to compare against the target of hate speech (10, 55). While it is not impossible for this classification to be correct, we also note the possibility that this is a case where the weaknesses of using text embeddings from a psychiatric disorder community instead of individuals with confirmed diagnoses could be made apparent. In particular, individuals with Antisocial Personality Disorder may have misconstrued their lack of empathy for a lack of interest in social interaction. Comparison of the portions of space occupied by both communities on the TDA mapping shows a striking degree of overlap; Schizoid Personality Disorder is encompassing all of Antisocial Personality Disorder's space, though it covers some additional space as well. This could also explain why Antisocial Personality Disorder had a hate speech classification count that could be considered to be astonishingly low; the Schizoid Personality Disorder and Narcissistic Personality Disorder embeddings overlapped with its space, and the model ultimately selected those two over Antisocial Personality Disorder, as can be seen by the trends in Supplemental Figures N-P. However, this is not guaranteed to be the case; it really could be that Schizoid Personality Disorder has a predilection for hate speech. Observation of another Cluster A personality disorder, Schizotypal Personality Disorder, showed that its region of space maintained a degree of continuity with that of Schizoid Personality Disorder (Figure 6b, Supplemental Figure K), offering an argument that the Schizoid Personality Disorder community on Reddit actually could represent their namesake. Determining if there is as strong a connection between Schizoid Personality Disorder and hate speech as depicted by our zero-shot classification results will likely require textual embeddings gathered from individuals with confirmed diagnoses of Schizoid Personality Disorder.

Complex Post-Traumatic Stress Disorder had notable results when juxtaposed to PTSD. Once other disorder communities occupying the connection point with hate speech were excluded from zero-shot classification, CPTSD was assigned a substantial amount of hate speech embeddings, but PTSD did not receive any (Supplemental Figure P). Upon subsequent exclusion of CPTSD, the hate speech embeddings assigned to CPTSD did not transfer to PTSD, but instead to Borderline Personality Disorder (Supplemental Figure Q). This allows for several observations. First, speech patterns in Borderline Personality Disorder are more similar to those seen in hate speech embeddings assigned to CPTSD than PTSD is. Second, speech patterns in CPTSD are more similar to hate speech than those seen in PTSD. Third, the portion of space occupied by CPTSD has more overlap with the space occupied by Schizoid Personality Disorder and the Cluster B personality disorders (Figures 6a-d, Figure 7). All of these seem to indicate that, for the purposes of this set of embeddings, the CPTSD community on Reddit exhibits patterns much more reminiscent of Cluster B than PTSD, to the point that not one embedding assigned to CPTSD was reassigned to PTSD. In terms of the debate surrounding whether CPTSD is unique as a construct, this would suggest that CPTSD is distinct from PTSD, but not from Cluster B personality disorders, especially Borderline Personality Disorder, the other construct that has been proposed as being the same as

CPTSD (49,50). However, we only had one community each for CPTSD and PTSD; a larger number of communities would be better for establishing a proper spatial relationship between CPTSD, PTSD, and Borderline Personality Disorder.

Zero-shot classification of misinformation communities indicated that their speech patterns could be similar to those of anxiety disorders in some instances. We believe this effect to have been at least partially influenced by our use of communities related to COVID, a topic prone to inciting anxiety (57). Future studies into misinformation and psychiatric disorders may be best performed with embeddings from communities pertaining to a more diverse set of misinformation topics, and with a larger set of communities overall. Overall, our pure misinformation communities appeared to have far less psychiatric disorder classifications compared to communities dedicated to hate speech and their related misinformation, and they exhibited the characteristics anxiety disorders instead of the characteristics of Schizoid Personality Disorder or Cluster B Personality Disorders that were seen in hate speech communities. Despite this, most of the misinformation embeddings maintained a connection with the hate speech region of space, and by extension the psychiatric disorder space. This suggests that there is a connection between psychiatric disorders and misinformation, though our dataset for this study is insufficient for properly comprehending it.

The greatest limitation of our study is that it must be noted that our results are not based on confirmed diagnoses; firstly, we performed classifications of embeddings that comprised an aggregate of users online, and psychiatric diagnoses are attributed to individuals, not groups of people. Secondly, our training data for psychiatric disorders is sourced from individuals claiming to have a particular psychiatric disorder; we have no evidence as to whether or not they speak the truth. However, several portions of our results indicate that these psychiatric disorder communities functioned as effective proxies. First, the high accuracy of our model when using distilled embeddings as testing data indicates that the model really does identify embeddings from psychiatric disorder communities on Reddit to be different from those of our control communities. Combined with the most prominent inaccuracies in Figure 1 being ones where a psychiatric disorder community was misclassified as a community for a psychiatric disorder with strong ties to it, this would appear to suggest that these embeddings of psychiatric communities served as substantially effective proxies for their respective psychiatric disorders. Second, observation of our TDA mapping shows that all of our psychiatric disorder communities shared the same region of space; that is to say, 866 embeddings comprising over 800,000 posts taken from 34 different psychiatric disorder communities were analyzed, and not one of them was considered as being so different from the others that they lacked an edge connecting them to the psychiatric disorder region of the mapping. Juxtaposing this to the control embeddings' general pattern of occupying their own individualized regions of space shows that this set of psychiatric disorder embeddings is very unlikely to involve communities that do not represent at least some form of psychiatric disorder. Consider further how these disorder communities demonstrate overlap similarly to how one would expect their comorbidities would be (e.g. ADHD having a wide range over the space, signifying overlap with many disorders, Cluster B disorders sharing the same region of space, depression overlapping most of the suicidality space), and it becomes apparent that even if these communities are serving as proxies for these disorders, they are serving fairly well, with the potential exception of Schizoid Personality Disorder.

We had several other limitations in this study; time/cost limitations meant that our sample size of posts for our embeddings were somewhat smaller than ideal. Additionally, we do not have every psychiatric disorder represented here; notably, we do not have Histrionic Personality Disorder, one of the Cluster B personality disorders. Furthermore, Reddit was the only social media site we drew from; other sites may have different user demographics, and potentially different results.

Our use of what we defined as distilled embeddings as training data for classification tasks yielded unexpected results. While they proved superior to IUP embeddings when it came to classifying embeddings of their own kind, they were inferior in performance for zero-shot classification or when the testing data was comprised of IUP embeddings. It is possible that our method of aggregating multiple posts together caused some components of the speech patterns for a community to be effectively cancelled out inside the aggregated embedding, impairing their ability to properly classify. However, models trained on IUP embeddings much more accurately classified a distilled embedding testing dataset than one composed of IUP embeddings, suggesting that even with some speech patterns lost, distilled embeddings were better representatives of their communities (and therefore their proposed psychiatric disorder or some other construct) than IUP embeddings. We therefore believe that distilled embeddings have some value if used as representatives of a construct in a testing dataset whilst using a model trained on IUP embeddings. However, an alternative explanation for the witnessed inaccuracies may lie in the overclassification of substance use disorders. The total number of distilled embeddings categorized as representing substance use disorder communities is 111, which is by far the highest number of embeddings assigned to any psychiatric disorder category. There is the possibility that this caused the model to become overly sensitive when it comes to detecting substance use disorders; future studies with more balanced numbers of distilled embeddings for each psychiatric disorder may be able to avoid this sensitivity issue.

We believe it is important to emphasize that these results do not necessarily indicate that individuals with psychiatric disorders will engage in hate speech or misinformation, and they most certainly do not indicate that they are the primary source of these issues. The matter of the directionality of the relationship is not established, either: Is the relationship because those with psychiatric disorders are more vulnerable to hate speech, or is it that hate speech communities take on speech patterns resembling psychiatric disorders as they develop? Further studies may be able to better answer these questions.

## Conclusion

In conclusion, our results indicate that hate speech communities on Reddit share speech patterns with Reddit communities for Schizoid Personality Disorder, three Cluster B personality disorders (Borderline, Narcissistic, and Antisocial), and Complex Post-Traumatic Stress Disorder. While the Cluster B disorders would be expected due to prior studies into the Dark Triad and hate speech, this nonetheless offers confirmation that they are acting similarly to their Dark Triad counterparts when it comes to hate speech. Furthermore, the two non-Cluster B disorders have not been discussed in hate speech or misinformation literature to the best of our knowledge, and offer new routes of investigation.

While personality disorders are historically resistant to treatment, the successes in treating Borderline Personality Disorder indicate that treatment is not impossible (58,59). More importantly, the similarities displayed between our embeddings of hate speech and Cluster B personality disorder communities could potentially indicate that the principles of methods such as

Dialectical Behavior Therapy may have an effect on hate speech, though this is not necessarily guaranteed.

Misinformation's relations to psychiatric disorder communities proved to remain elusive; a clear connection can be seen in the TDA mapping, and zero-shot classification had over 25% of the misinformation embeddings assigned to psychiatric disorders, but it was difficult to elucidate a clear pattern with the dataset at hand. A more robust dataset may be able to identify clearer trends.

While Topological Data Analysis may run into computational challenges with a high number of IUP embeddings, the use of distilled embeddings allows for TDA to handle over a million posts in a single analysis. While their use is currently effectively limited to testing data, it may be possible to improve them to render them usable for training data as well.

Zero-shot classification proved to be an effective tool for identifying which psychiatric disorder communities were most similar to hate speech and misinformation communities. However, the addition of TDA allowed for a visualization of the space, and ultimately allowed for the observation that psychiatric disorders effectively had a spectrum of similarity when it came to hate speech. It also showed that all of our psychiatric disorder community embeddings were connected and similar to each other, and by extension, hate speech and misinformation, something that the vast majority of control embeddings failed to do. Up to this point, TDA has gone relatively unconsidered in psychiatric research; we suggest making use of its ability to visualize the space that one is using for classification tasks, as it can allow for the identification of routes of investigation that may have otherwise gone unnoticed.

## Data and Code Availability

The dataset and codes used in this study will be available at github.com/Andrew7178/TopologicalDataMappingOnlineHateSpeechMisinformationGeneralMentalHealth.

## Acknowledgement

We thank TAMU Academy of Physician Scientists for a Burroughs Wellcome Fund Scholarship to the first author.

**Supplemental Materials**

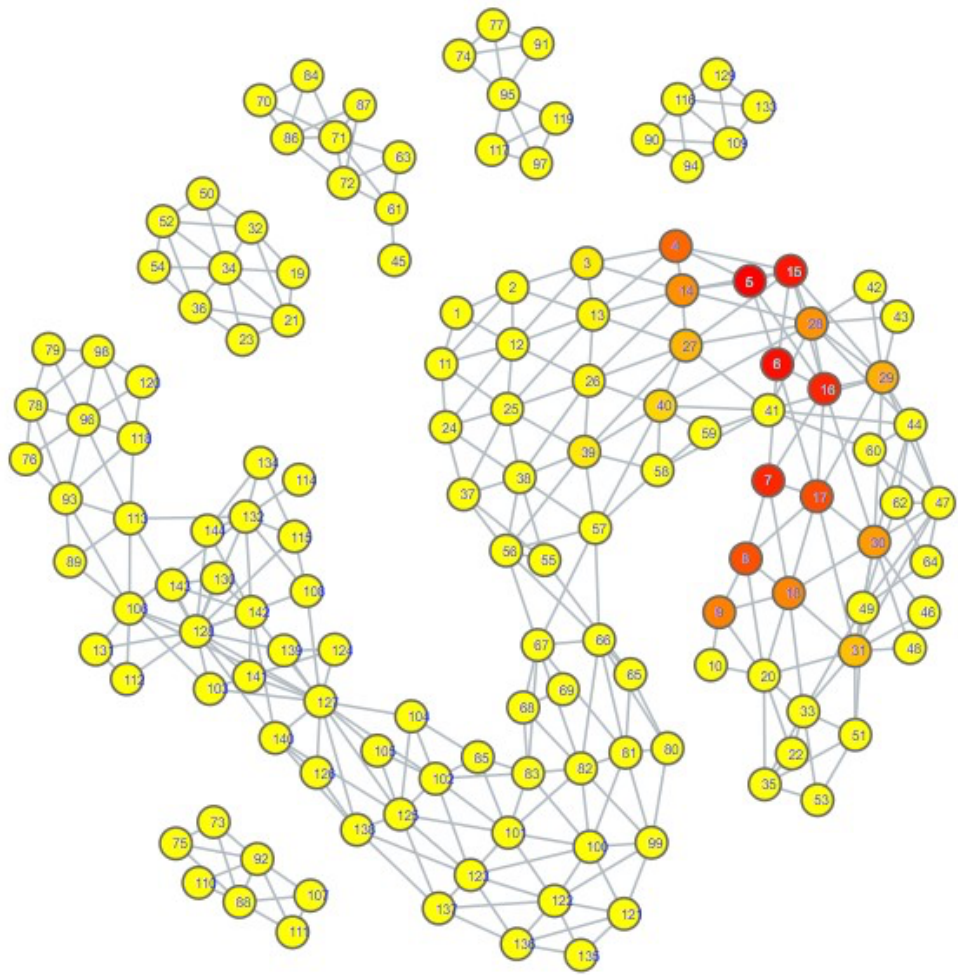

**Supplemental Figure A.** TDA Mapping of All Distilled Embeddings, Colored by Attention Deficit Hyperactivity Disorder Communities. Yellow = No Attention Deficit Hyperactivity Disorder Embeddings, Red = Entirely Comprised of Attention Deficit Hyperactivity Disorder Embeddings.

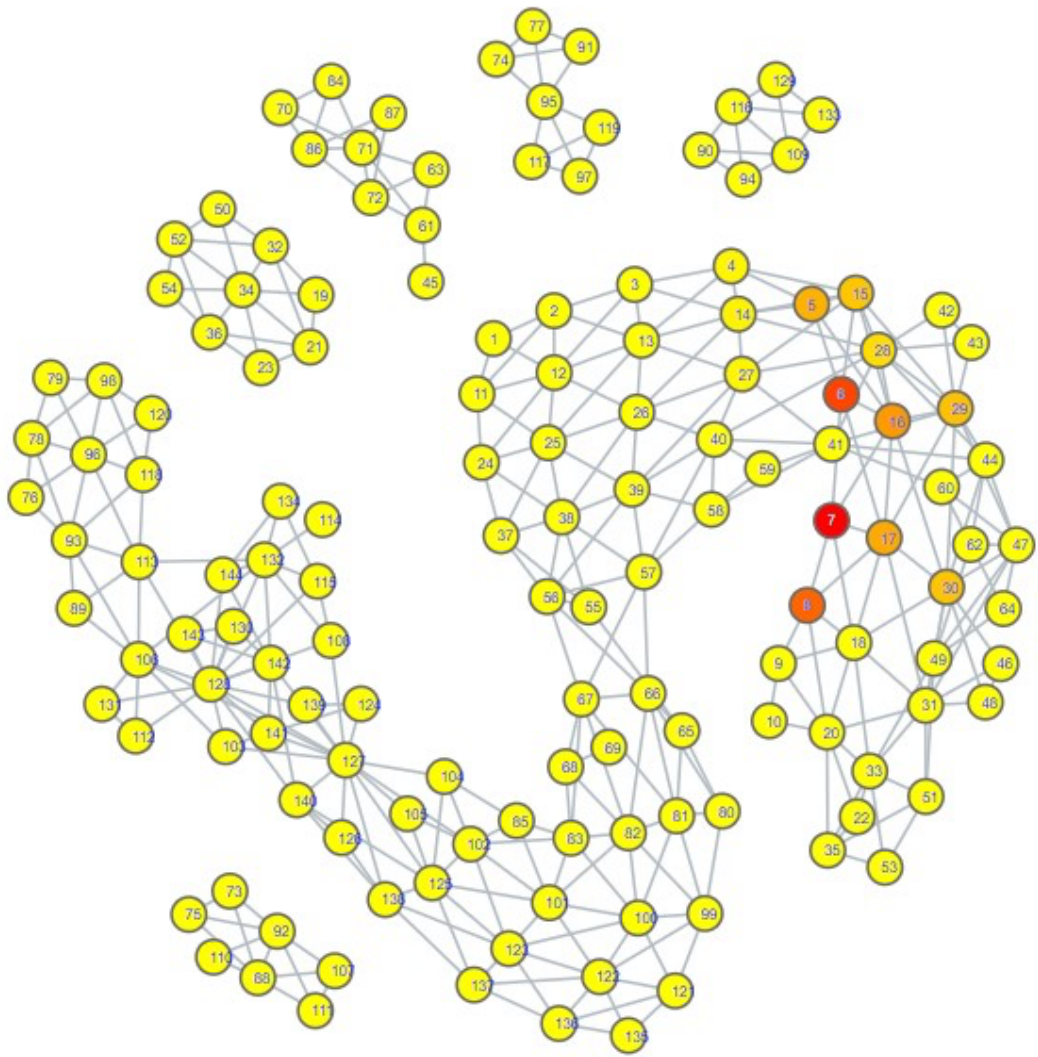

**Supplemental Figure B.** TDA Mapping of All Distilled Embeddings, Colored by Anxiety Disorder Communities. Yellow = No Anxiety Disorder Embeddings, Red = Entirely Comprised of Anxiety Disorder Embeddings.

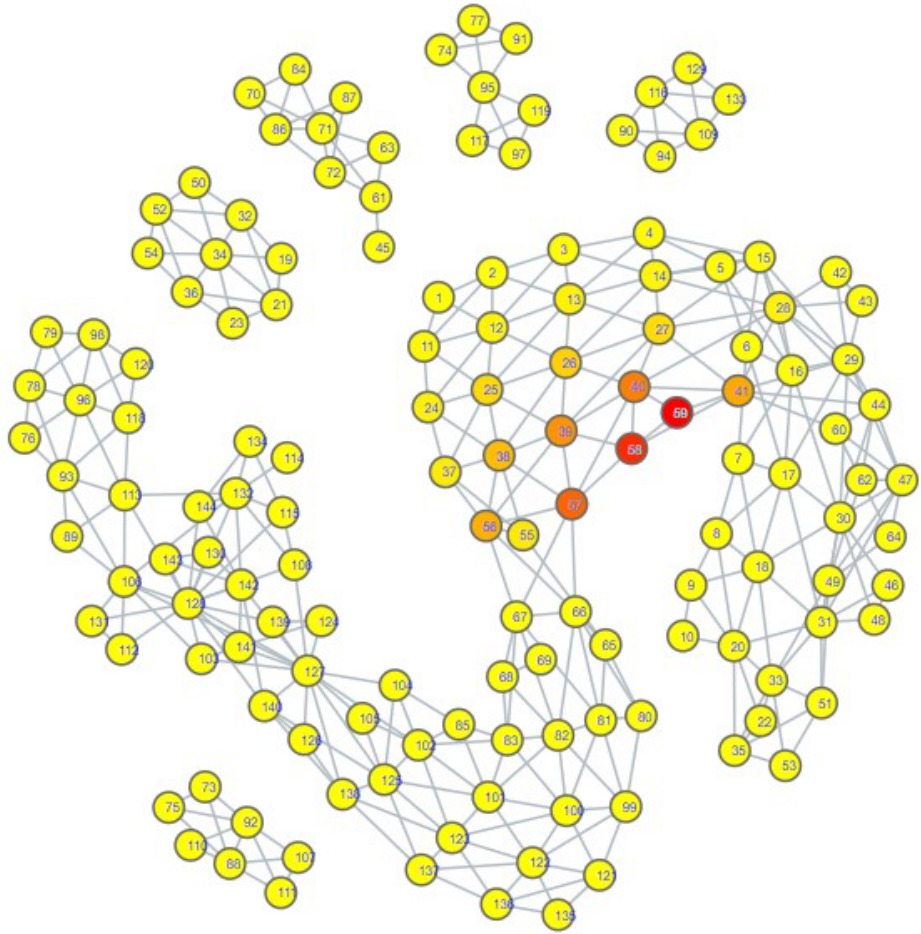

**Supplemental Figure C.** TDA Mapping of All Distilled Embeddings, Colored by Autism Spectrum Disorder Communities. Yellow = No Autism Spectrum Disorder Embeddings, Red = Entirely Comprised of Autism Spectrum Disorder Embeddings.

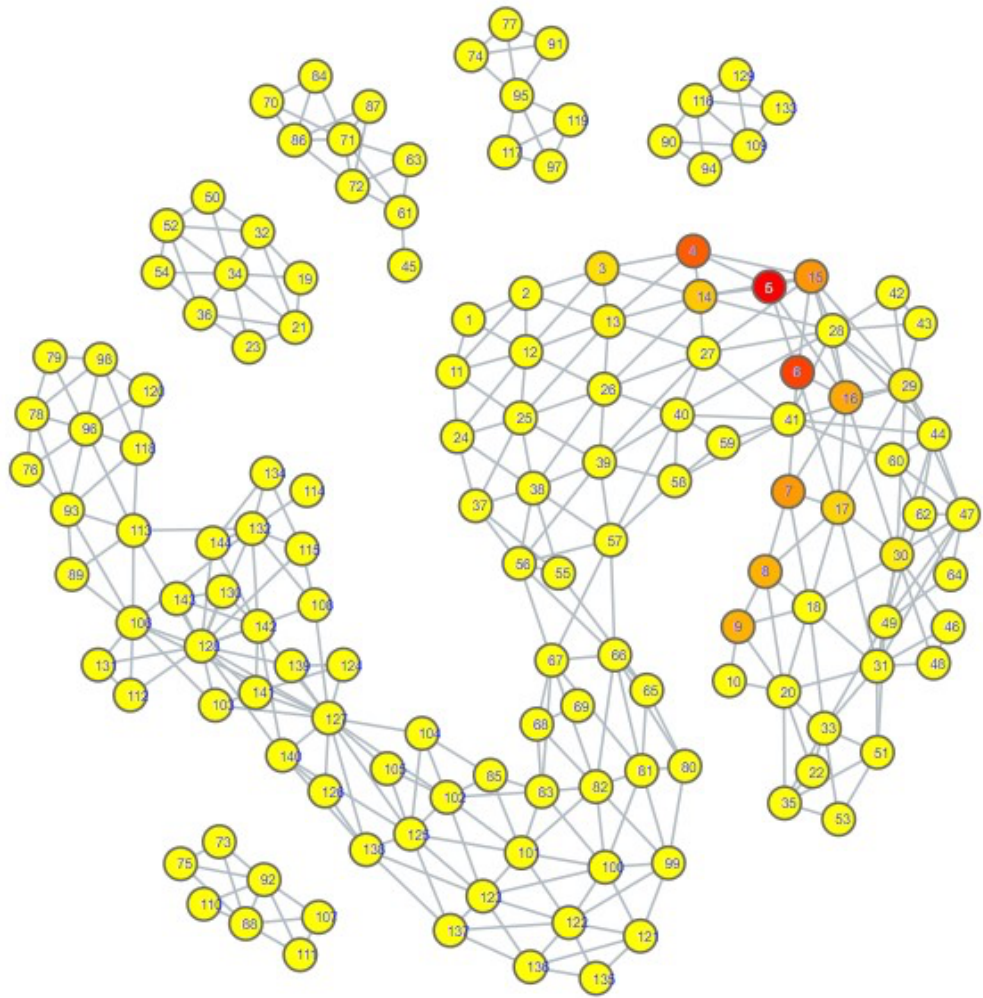

**Supplemental Figure D.** TDA Mapping of All Distilled Embeddings, Colored by Bipolar Disorder Communities. Yellow = No Bipolar Disorder Embeddings, Red = Entirely Comprised of Bipolar Disorder Embeddings.

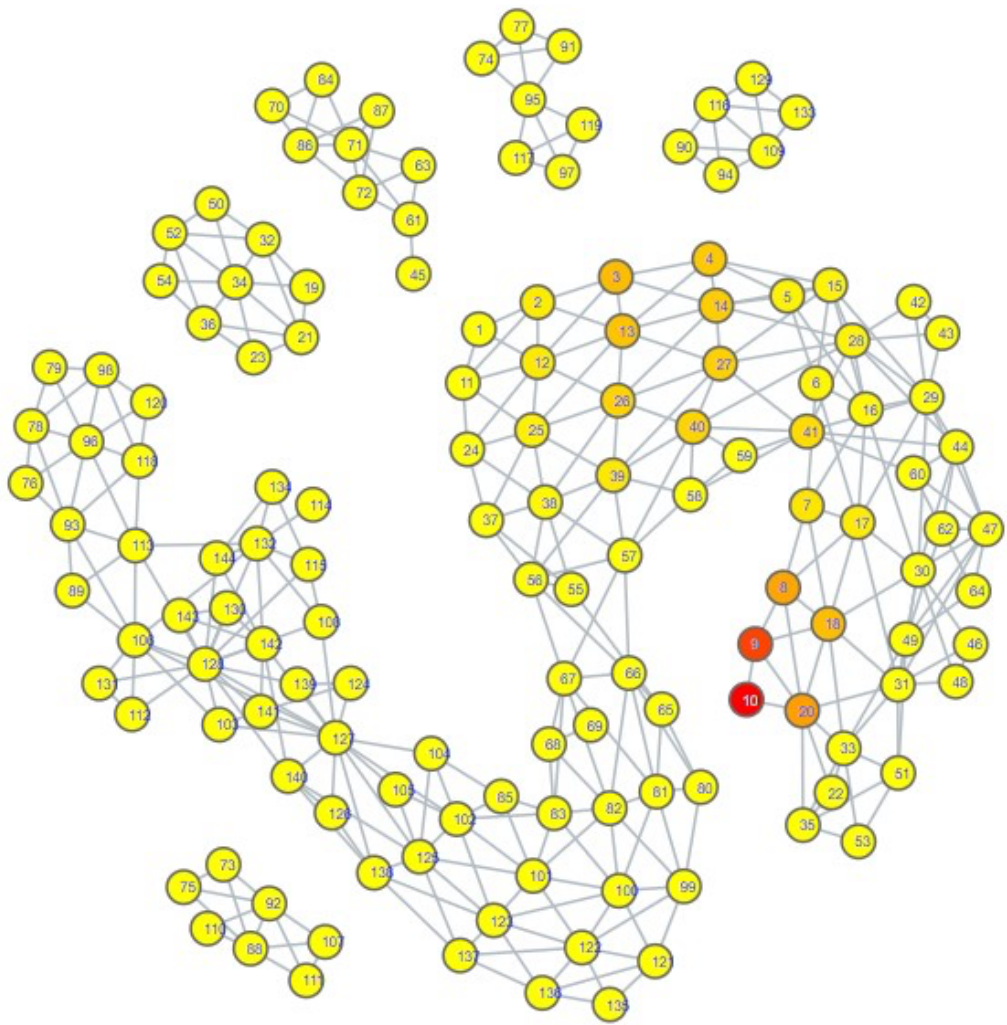

**Supplemental Figure E.** TDA Mapping of All Distilled Embeddings, Colored by Depression Communities. Yellow = No Depression Embeddings, Red = Entirely Comprised of Depression Embeddings.

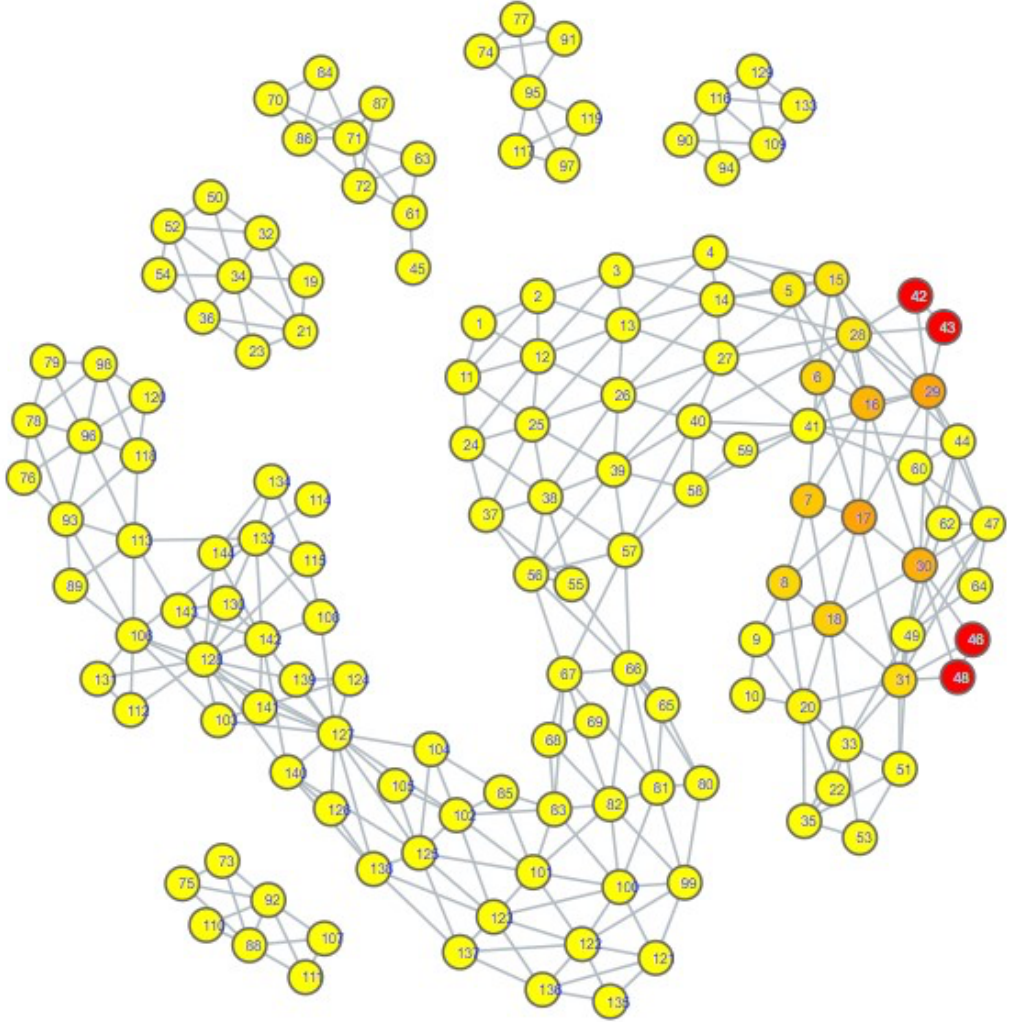

**Supplemental Figure F.** TDA Mapping of All Distilled Embeddings, Colored by Eating Disorder Communities. Yellow = No Eating Disorder Embeddings, Red = Entirely Comprised of Eating Disorder Embeddings.

**Supplemental Figure G.** TDA Mapping of All Distilled Embeddings, Colored by Obsessive Compulsive Disorder Communities. Yellow = No Obsessive Compulsive Disorder Embeddings, Red = Entirely Comprised of Obsessive Compulsive Disorder Embeddings.

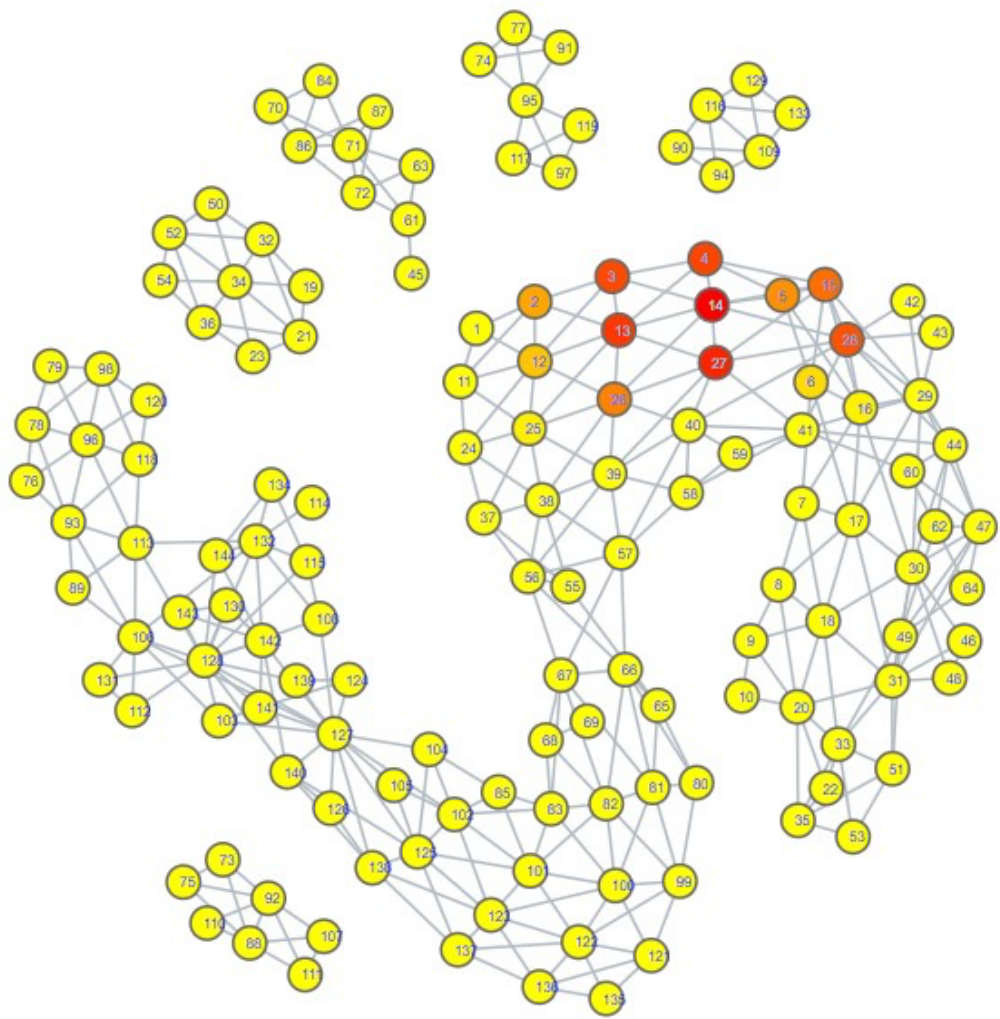

**Supplemental Figure H.** TDA Mapping of All Distilled Embeddings, Colored by Post-Traumatic Stress Disorder Communities. Yellow = No Post-Traumatic Stress Disorder Embeddings, Red = Entirely Comprised of Post-Traumatic Stress Disorder Embeddings.

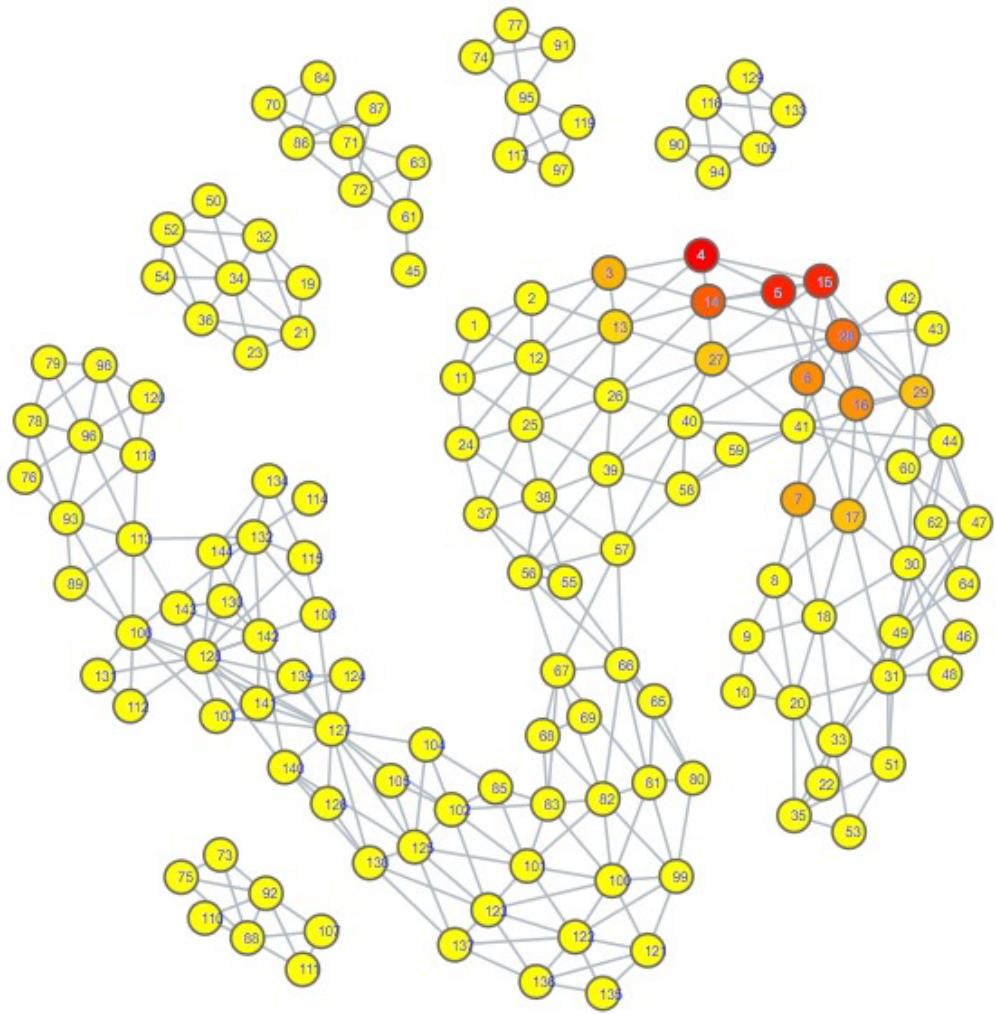

**Supplemental Figure I.** TDA Mapping of All Distilled Embeddings, Colored by Schizoaffective Disorder Communities. Yellow = No Schizoaffective Disorder Embeddings, Red = Entirely Comprised of Schizoaffective Disorder Embeddings.

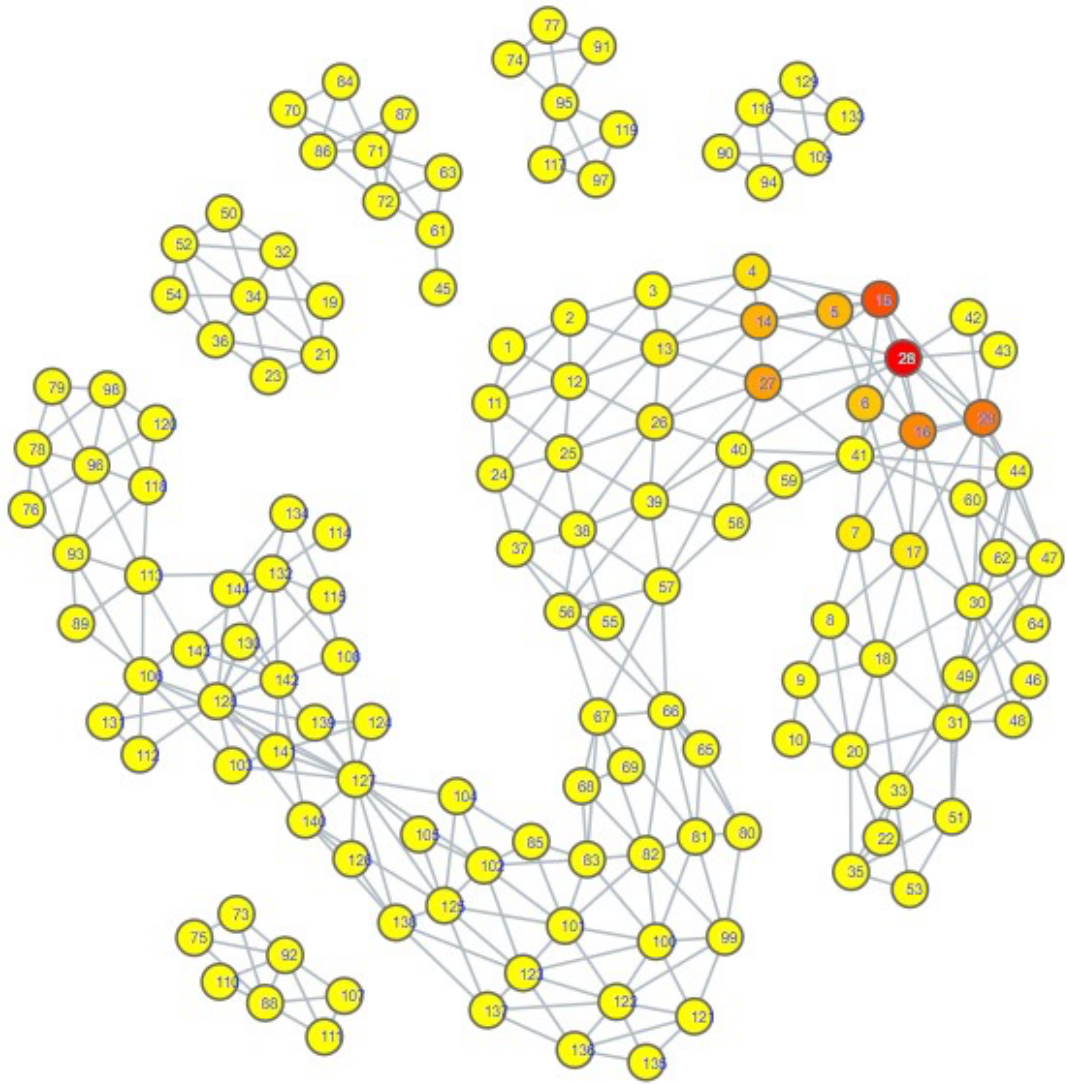

**Supplemental Figure J.** TDA Mapping of All Distilled Embeddings, Colored by Schizophrenia Communities. Yellow = No Schizophrenia Embeddings, Red = Entirely Comprised of Schizophrenia Embeddings.

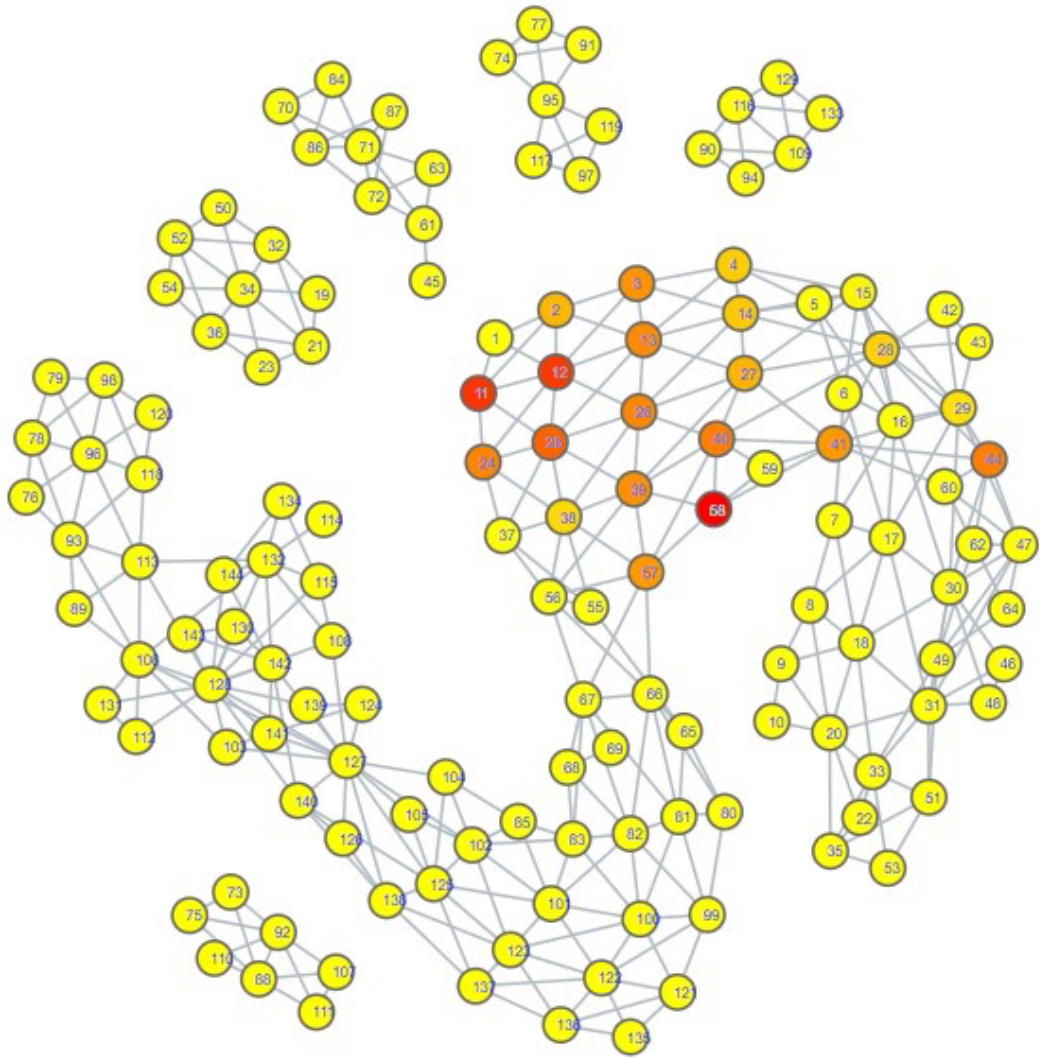

**Supplemental Figure K.** Topological Mapping Colored by Schizotypal Personality Disorder Communities. Yellow = No Schizotypal Personality Disorder Embeddings, Red = Entirely Comprised of Schizotypal Personality Disorder Embeddings.

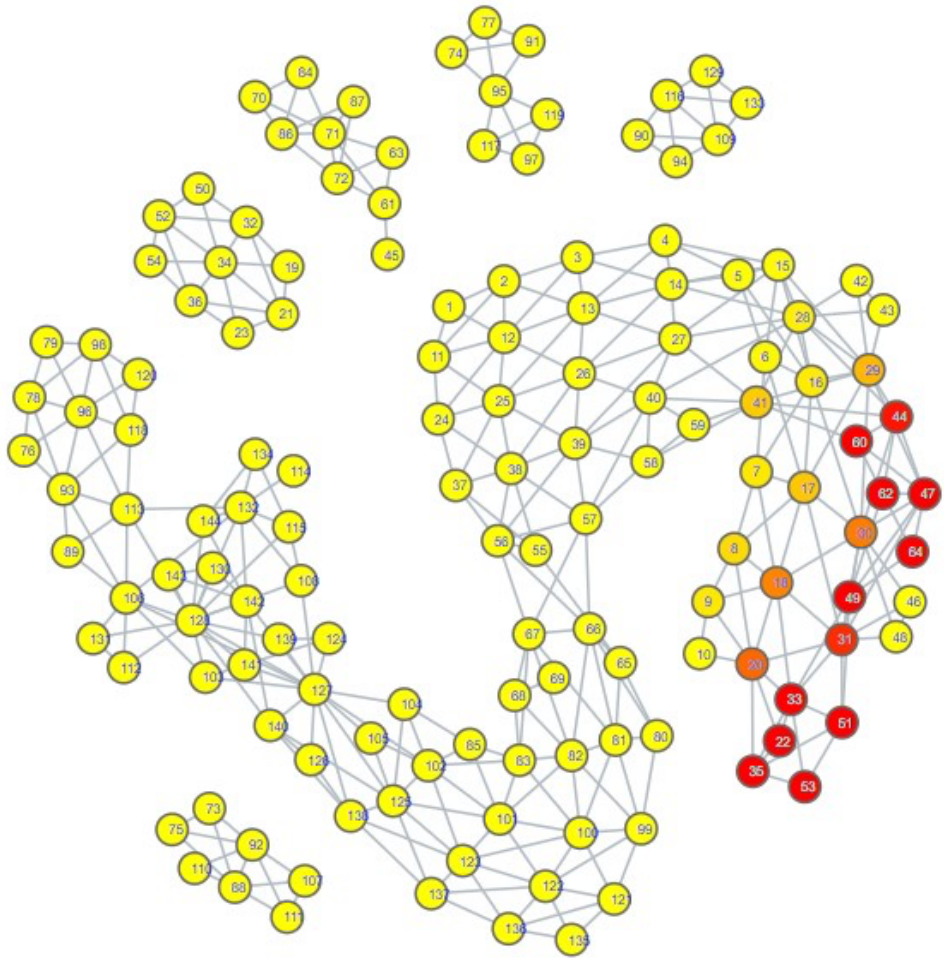

**Supplemental Figure L.** Topological Mapping Colored by Substance Use Disorder Communities. Yellow = No Substance Use Disorder Embeddings, Red = Entirely Comprised of Substance Use Disorder Embeddings.

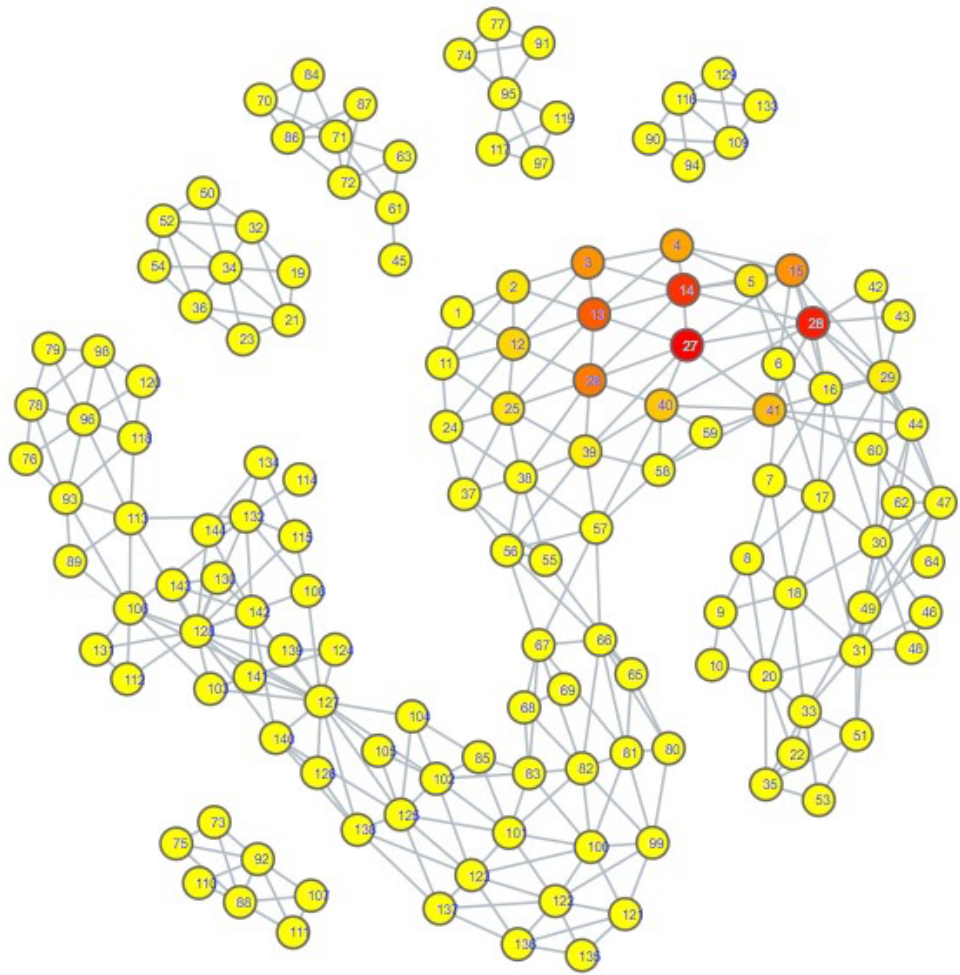

**Supplemental Figure M.** TDA Mapping of Suicidality Communities Pertaining to Suicide. Yellow = No Suicidality Embeddings, Red = Entirely Comprised of Suicidality Embeddings.

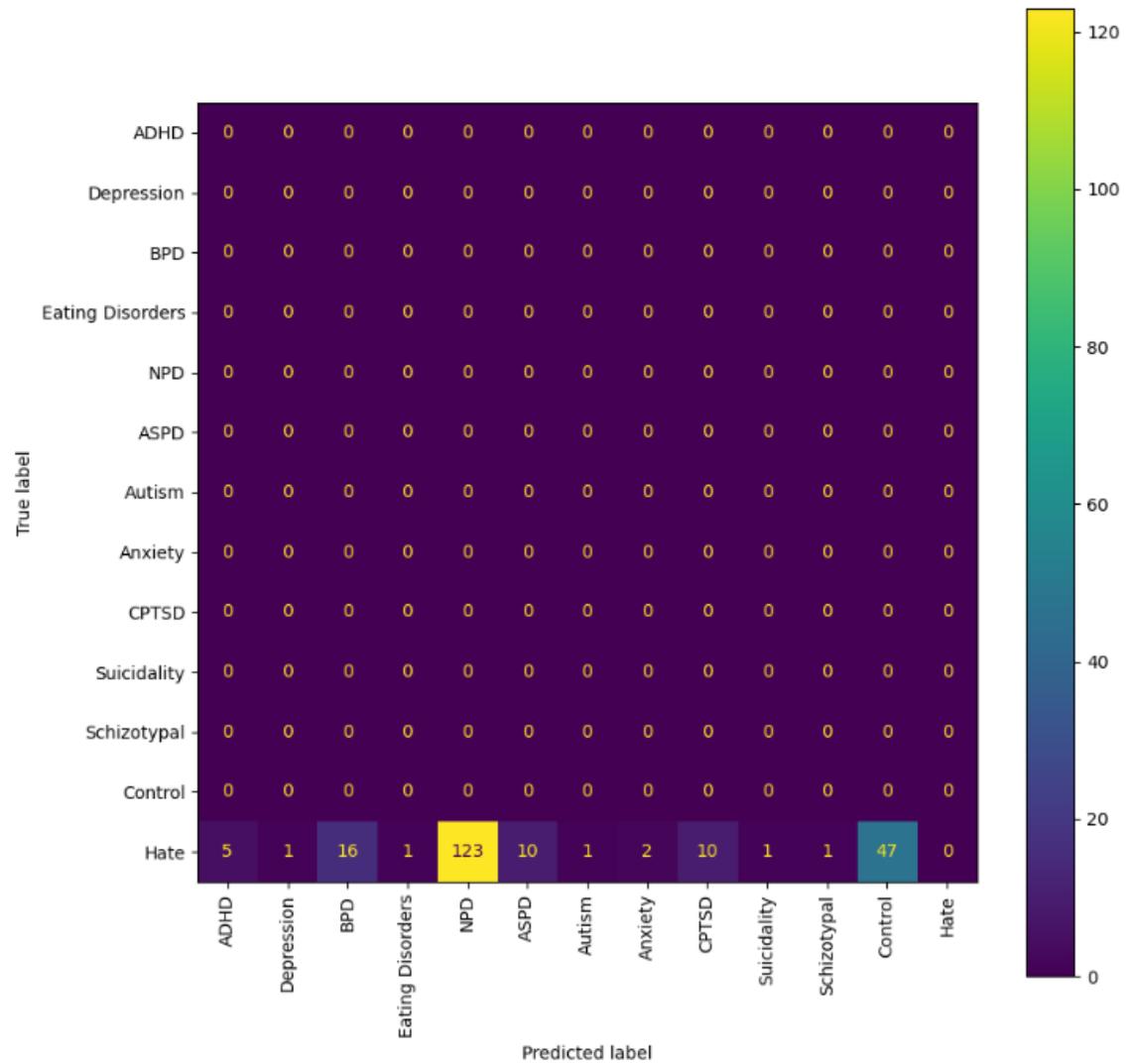

**Supplemental Figure N.** Zero-Shot Classification of Hate Speech, Excluding Schizoid Personality Disorder.

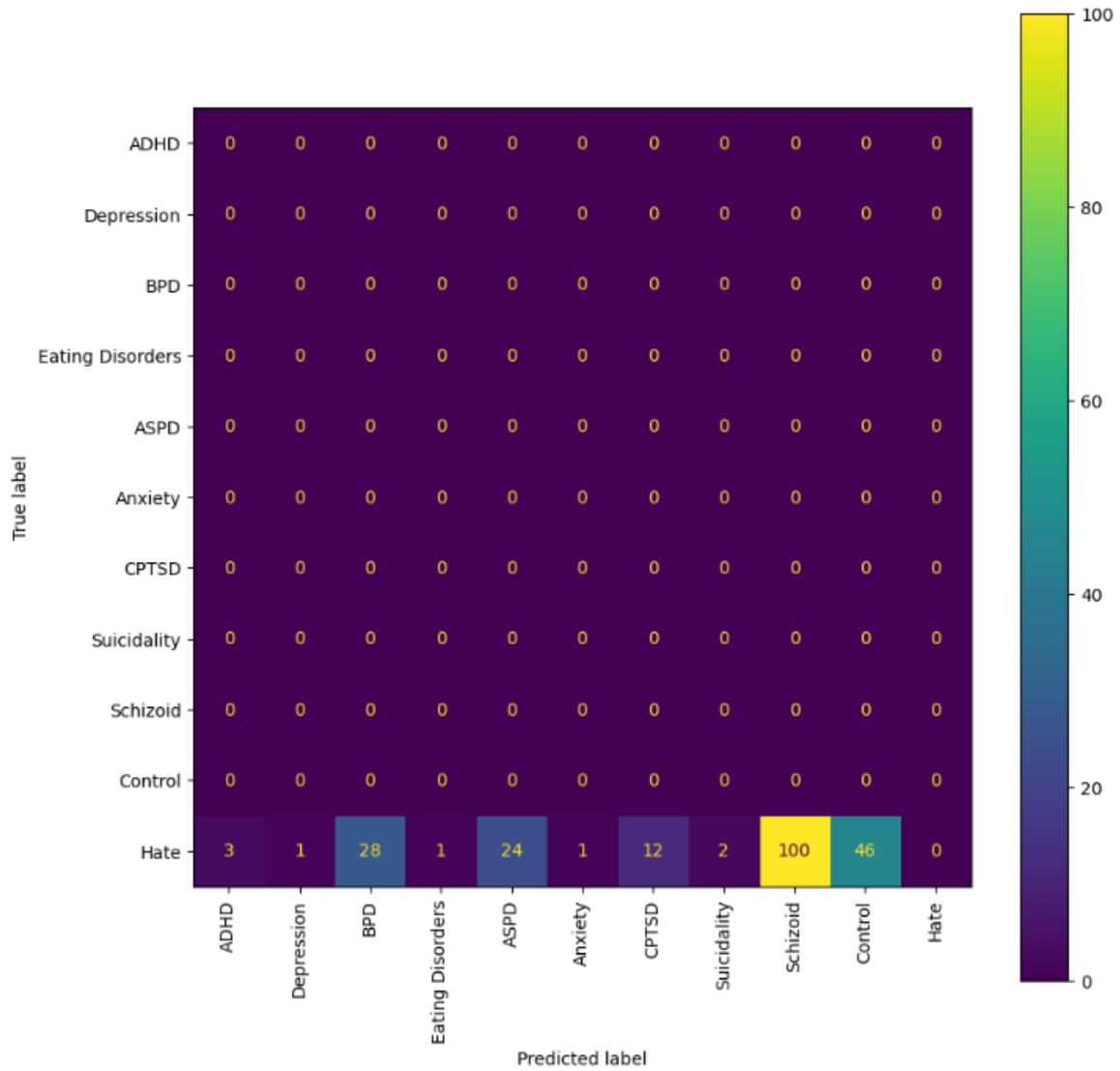

**Supplemental Figure O.** Zero-Shot Classification of Hate Speech, Excluding Narcissistic Personality Disorder.

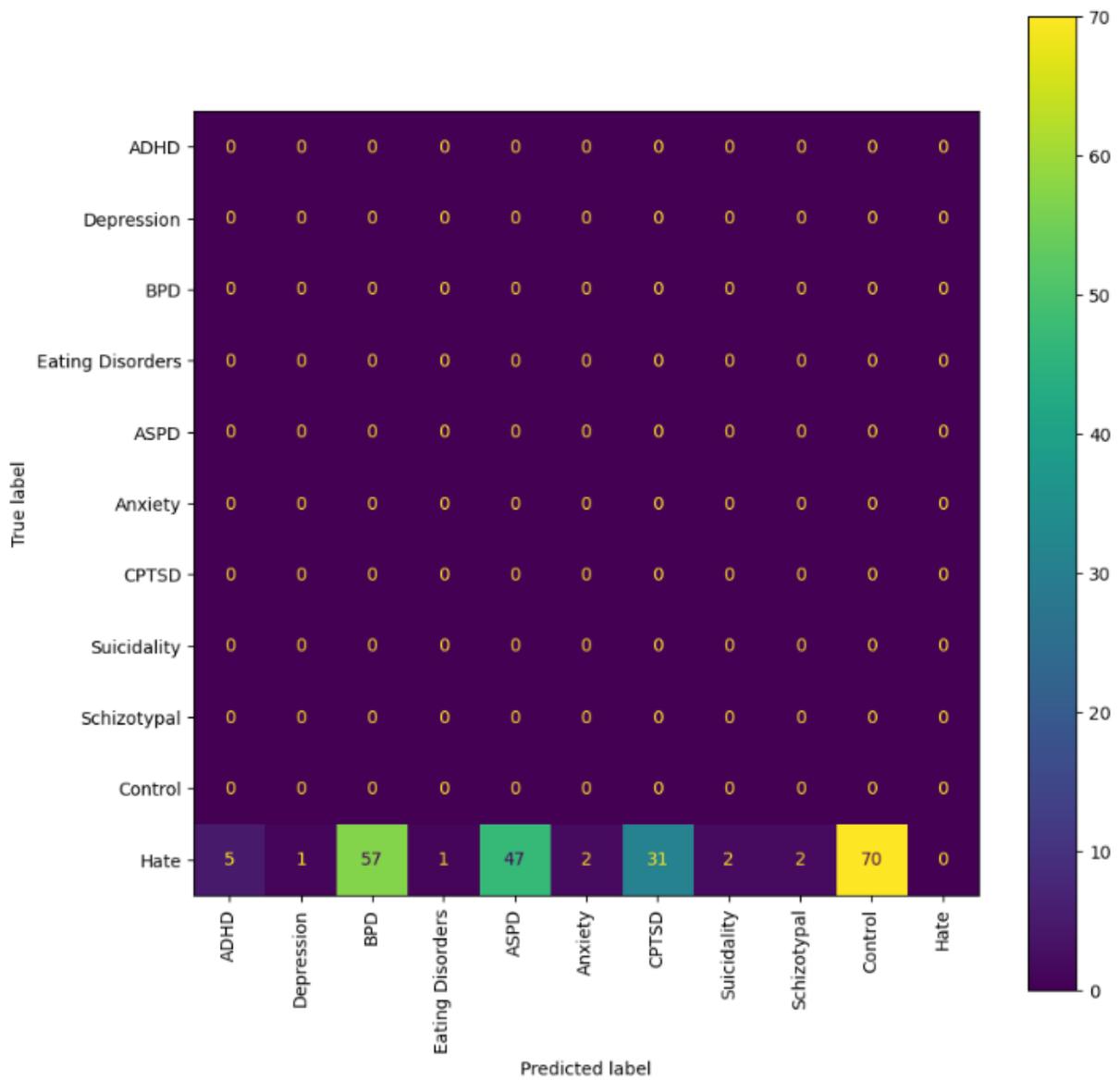

**Supplemental Figure P.** Zero-Shot Classification of Hate Speech, Excluding Schizoid Personality Disorder and Narcissistic Personality Disorder.

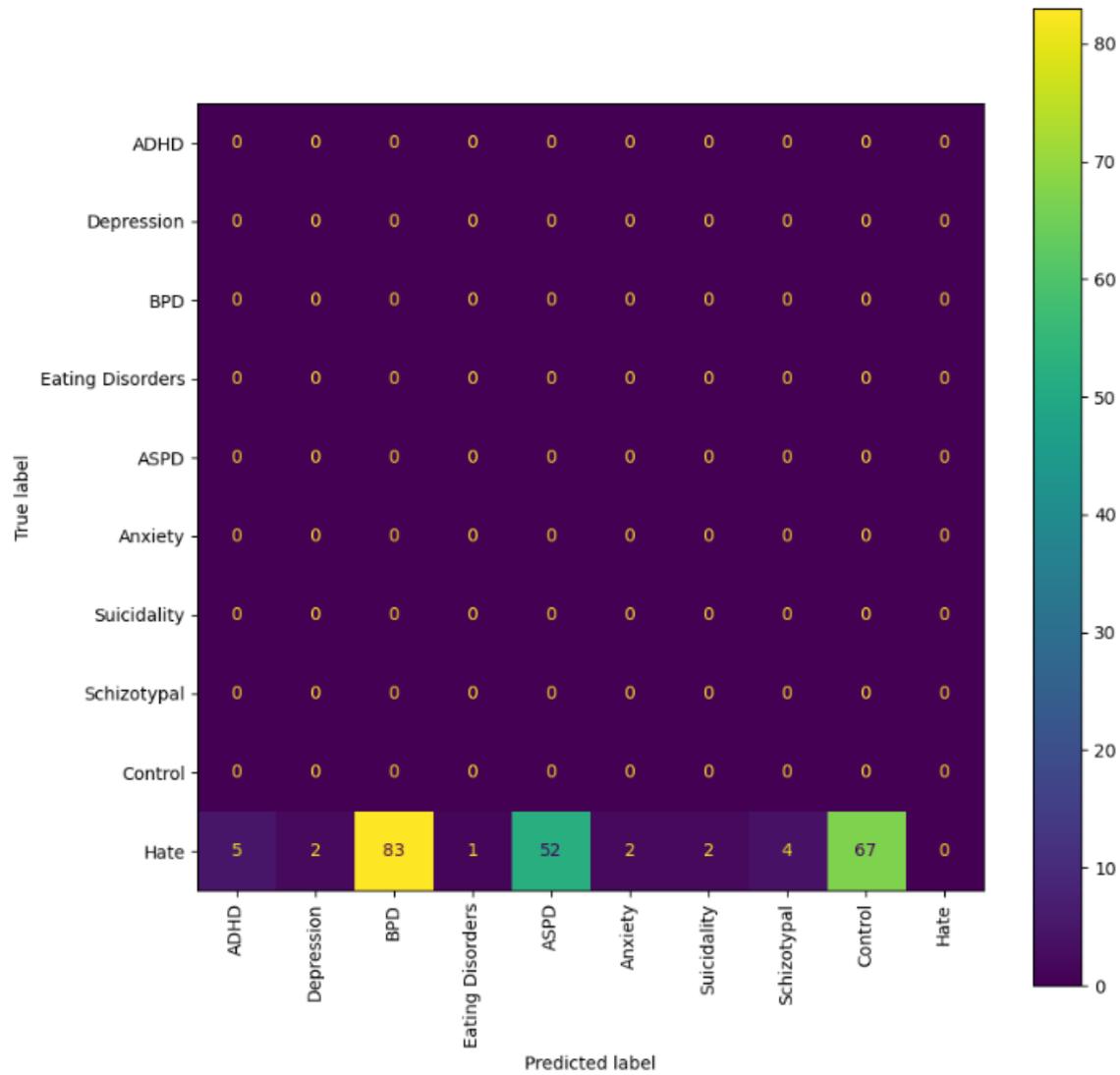

**Supplemental Figure Q.** Zero-Shot Classification of Hate Speech, Excluding Schizoid Personality Disorder, Narcissistic Personality Disorder, and Complex Post-Traumatic Stress Disorder.